\documentclass{article}

\usepackage{microtype}
\usepackage{graphicx}
\usepackage{subfigure}
\usepackage{booktabs} % for professional tables

\usepackage{hyperref}

\usepackage[accepted]{icml2024}

\usepackage{amsmath}
\usepackage{amssymb}
\usepackage{mathtools}
\usepackage{amsthm}

\usepackage[capitalize,noabbrev]{cleveref}

\theoremstyle{plain}

\theoremstyle{definition}

\theoremstyle{remark}

\usepackage[textsize=tiny]{todonotes}

\usepackage{xfrac}
\usepackage{dsfont}
\DeclareMathOperator*{\argmax}{arg\,max}

\usepackage{wrapfig}
\usepackage{xcolor}
\usepackage{nicematrix}
\usepackage{makecell}
\usepackage{colortbl}
\definecolor{lightblue}{rgb}{0.8,0.9,1}
\usepackage{titletoc}
\newcommand\DoToC{%
  \startcontents
  \printcontents{}{2}{\textbf{Contents}\vskip3pt\hrule\vskip5pt}
  \vskip3pt\hrule\vskip5pt
}
\icmltitlerunning{Enhanced Utility and Robustness in Smoothed DRL Agents}

\begin{document}

\twocolumn[
\icmltitle{Breaking the Barrier: \\ Enhanced Utility and Robustness in Smoothed DRL Agents}




\begin{icmlauthorlist}
\icmlauthor{Chung-En Sun}{ucsd}
\icmlauthor{Sicun Gao}{ucsd}
\icmlauthor{Tsui-Wei Weng}{ucsd}
\end{icmlauthorlist}

\icmlaffiliation{ucsd}{UC San Diego}
\icmlcorrespondingauthor{Chung-En Sun}{cesun@ucsd.edu}
\icmlcorrespondingauthor{Tsui-Wei Weng}{lweng@ucsd.edu}

\icmlkeywords{Machine Learning, ICML}

\vskip 0.3in
]



\printAffiliationsAndNotice{}  % leave blank if no need to mention equal contribution

\begin{abstract}
Robustness remains a paramount concern in deep reinforcement learning (DRL), with randomized smoothing emerging as a key technique for enhancing this attribute. However, a notable gap exists in the performance of current smoothed DRL agents, often characterized by significantly low clean rewards and weak robustness. In response to this challenge, our study introduces innovative algorithms aimed at training effective smoothed robust DRL agents. We propose S-DQN and S-PPO, novel approaches that demonstrate remarkable improvements in clean rewards, empirical robustness, and robustness guarantee across standard RL benchmarks. Notably, our S-DQN and S-PPO agents not only significantly outperform existing smoothed agents by an average factor of $2.16\times$ under the strongest attack, but also surpass previous robustly-trained agents by an average factor of $2.13\times$. This represents a significant leap forward in the field. Furthermore, we introduce Smoothed Attack, which is $1.89\times$ more effective in decreasing the rewards of smoothed agents than existing adversarial attacks. Our code is available at: \href{https://github.com/Trustworthy-ML-Lab/Robust_HighUtil_Smoothed_DRL}{https://github.com/Trustworthy-ML-Lab/Robust\_HighUtil\_Smoothed\_DRL}
\end{abstract}

\section{Introduction}
\label{sec:introduction}

\begin{figure*}[!t]
\centering
\includegraphics[width=0.95\textwidth]{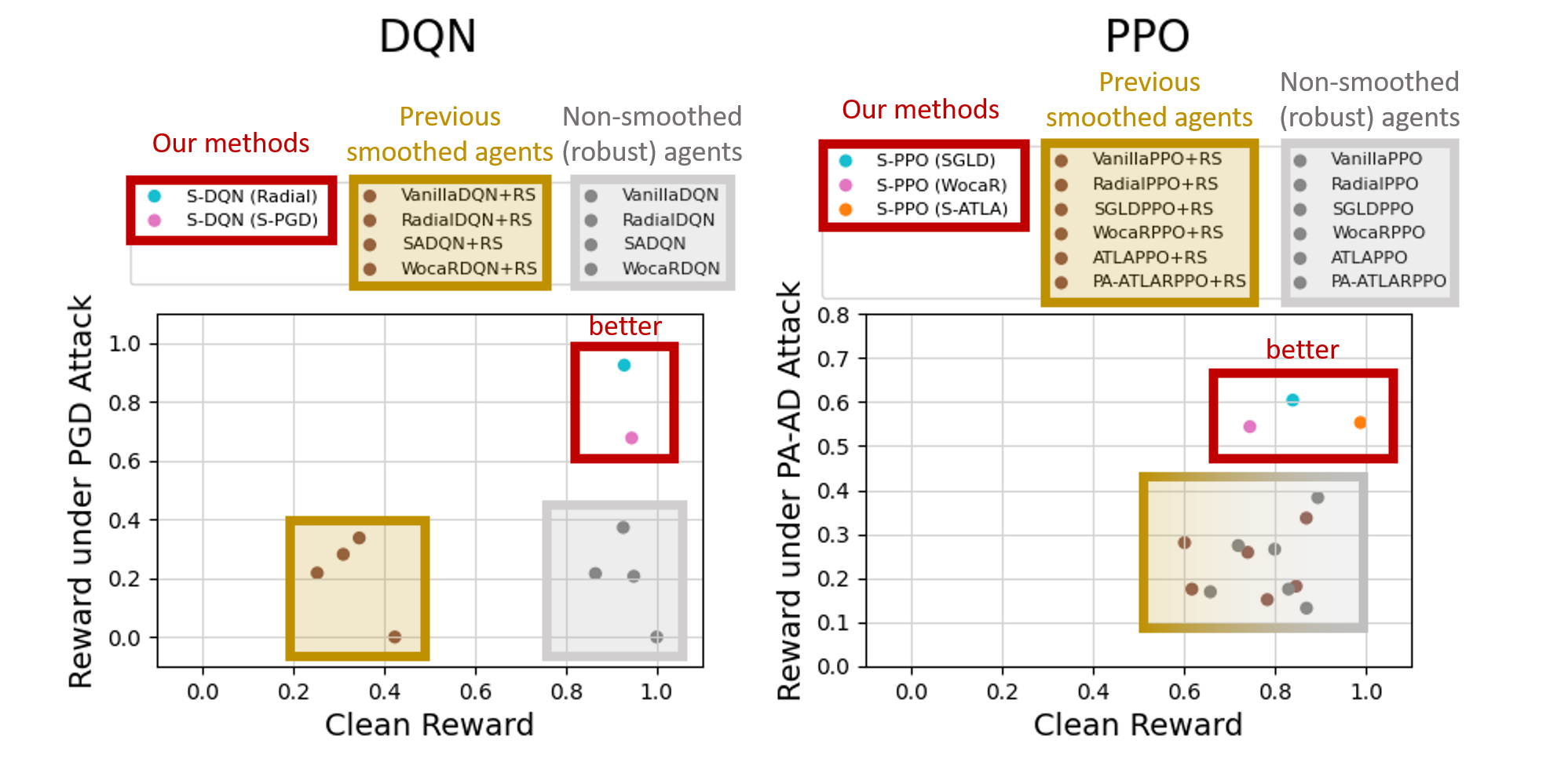}
\vspace{-20pt}
\caption{The clean reward and reward under attack for DQN and PPO agents. The presented reward is normalized and averaged across environments. Our S-DQN and S-PPO agents (in the \textcolor{red}{Red} boxes) exhibit significantly improved clean reward and robustness in comparison to the previous smoothed agents (in the \textcolor{brown}{Brown} boxes) and the non-smoothed robust agents (in the \textcolor{gray}{Gray} boxes).}
\label{fig:trade-off}
\end{figure*}

Deep Reinforcement Learning (DRL) has achieved remarkable performance, surpassing human-level capabilities in various game environments \cite{atari, go}.
%, as well as in safety-critical domains such as robotics \cite{robotics1, robotics2}, autonomous driving \cite{driving}, and healthcare \cite{healthcare}. 
However, recent studies have unveiled a significant vulnerability within DRL -- its susceptibility to adversarial perturbations \cite{Huang, Lin, Weng}. As a result, it is imperative to enhance the robustness of DRL agents before deploying them in real-world applications, especially those involving safety-critical tasks.

In response to this need, researchers have adapted techniques from robust classifier training to bolster DRL agents' resilience  \cite{advrl, sa, radialrl}. This includes employing adversarial training strategies \cite{advrl} and introducing methods that enhance robustness through the use of robustness verification bounds \cite{sa,radialrl}. Additionally, a focus has shifted towards enabling certifiable robustness in DRL agents using Randomized Smoothing (RS) \cite{crop, policysmoothing}, transforming agents into their "smoothed" counterparts. However, this transformation traditionally occurs only during testing, without additional training.

% To address this challenge, techniques initially developed for training robust classifiers have been adapted to fortify the resilience of DRL agents. For instance, \cite{advrl} employed adversarial training \cite{PGD, FGSM} to train DRL agents against adversarial examples, while \cite{sa, radialrl} proposed methods to enhance DRL agents' robustness through regularizers based on robustness verification bounds \cite{IBP}.
%
%
% On the other hand, there is a vein of research to improve the certifiable robustness of DRL agents \cite{crop,policysmoothing}. The key idea is to transform a DRL agent into a smoothed agent using Randomized Smoothing (RS) \cite{rs}. Importantly, these works do not involve any training on the smoothed agents, as the transformation only involves applying RS during testing.

Unfortunately, despite the progress in enhancing DRL robustness, we found that existing smoothed agents \cite{crop, policysmoothing} demonstrate a notable deficiency: they yield substantially lower clean reward and show little improvement in robustness compared to their non-smoothed counterparts. This critical gap, which we will discuss in Section \ref{sec:trade off} "\textit{Failure in existing smoothed DRL agents}", has been largely overlooked in previous research. This highlights the need for more effective strategies. Furthermore, previous attack evaluations are ineffective at reducing the rewards of smoothed agents as discussed in section \ref{sec:sdqn} Table \ref{table:new attack}, potentially creating an illusion of empirical robustness.

Driven by these challenges, our work aims to significantly enhance the clean reward, robust reward, and robustness guarantee of smoothed DRL agents. We also address the overestimation of robustness in previous studies by introducing a novel smoothing strategy and a more effective attack method. As a result, we present two innovative agents, S-DQN and S-PPO, designed for both discrete and continuous action spaces. Our proposed agents not only achieve high clean rewards but also provide robustness certification, setting new state-of-the-art across various standard RL environments, including Atari games \cite{atari} and continuous control tasks \cite{gym}.

\break

Our contributions are two-fold:
\begin{enumerate}
    \item We identify and address the shortcomings in existing smoothed DRL agents, particularly concerning their low clean rewards and limited robustness. To address the limitation, we introduce the first robust DRL training algorithms utilizing Randomized Smoothing (RS) for both discrete actions (S-DQN) and continuous actions (S-PPO). Additionally, we introduce new smoothing strategies and a new attack (Smoothed Attack) to fix the overestimation of robustness in the previous works.
    \item Our agents establish a new state-of-the-art record on both robust reward and clean reward. Our S-DQN and S-PPO achieve a $2.52\times$ and $1.80\times$ increase in reward respectively, outperforming existing best \textbf{smoothed agents} under the strongest attack. Notably, our S-DQN and S-PPO also surpass previous \textbf{best (non-smoothed) robust agents} by $2.70\times$ and $1.58\times$ increase in reward respectively.
\end{enumerate}

We structure our paper as follows: In Section \ref{sec:trade off}, we discuss the issue of low clean reward in existing smoothed DRL agents. In Section \ref{sec:our approach}, we introduce the main algorithms of S-DQN and S-PPO. In Section, \ref{sec:certification}, we derive the robustness certification for S-DQN and S-PPO. In Section \ref{sec:experiment}, we evaluate the performance of S-DQN and S-PPO in terms of both robust reward and robustness guarantee. In Section \ref{sec:background}, we provide background information relevant to our work. Finally, in Section \ref{sec:conclusion}, we summarize our work and discuss potential future directions.
\section{Failure in existing Smoothed DRL Agents}
\label{sec:trade off}

% Despite randomized smoothing (RS) being amenable to providing robustness certification, we found that \textbf{the clean reward of all the previous smoothed agents is low} as we show in Figure \ref{fig:trade-off}. In the DQN setting, the clean reward of the previous smoothed DQN agents degrades significantly due to the noise introduced by RS, regardless the base DRL agents are robustly trained (RadialDQN, SADQN, WocaRDQN) or not (VanillaDQN). Meanwhile, the reward of the previous smoothed agents under PGD attack shows no improvement. A similar trend is evident in the PPO setting: smoothed PPO agents, such as SGLDPPO+RS \cite{sa} and VanillaPPO+RS, exhibit lower clean rewards compared to their non-smoothed counterparts, with only marginal improvement under attack scenarios.

Randomized Smoothing (RS) is a known technique for enhancing robustness in Deep Reinforcement Learning (DRL). However, our analysis reveals a critical drawback: \textbf{the clean reward of all previously studied smoothed agents is notably low with no improvement on the robust reward compared to the non-smoothed agents, as demonstrated with the yellow boxes in Figure \ref{fig:trade-off}}. In the DQN framework, previous smoothed agents experience notable reward degradation due to the noise from RS. This degradation persists even under attack scenarios, where no improvement in robust reward is observed. The same pattern is evident with PPO agents: the previous smoothed agents display diminished clean rewards compared to their non-smoothed versions, with only marginal enhancements on the robust reward. For further context on these previous studies, please refer to Section \ref{sec:background}.

In contrast, our proposed S-DQN and S-PPO, highlighted in Figure \ref{fig:trade-off} with red boxes, outperform all the previous smoothed agents \cite{crop, policysmoothing} and non-smoothed robust agents \cite{sa, radialrl, wocar, optimalattack, paad} in both robustness and clean reward. This suggests the feasibility of mitigating the adverse effects of randomized smoothing while significantly enhancing robustness. In the following section, we introduce our novel approaches: S-DQN for discrete actions and S-PPO for continuous actions.

% In stark contrast, as illustrated in Figure \ref{fig:trade-off}, our proposed methods (highlighted in red boxes) demonstrate the capacity to achieve both high robustness and clean reward.  

\begin{figure*}[ht]
\centering
\includegraphics[width=1\textwidth]{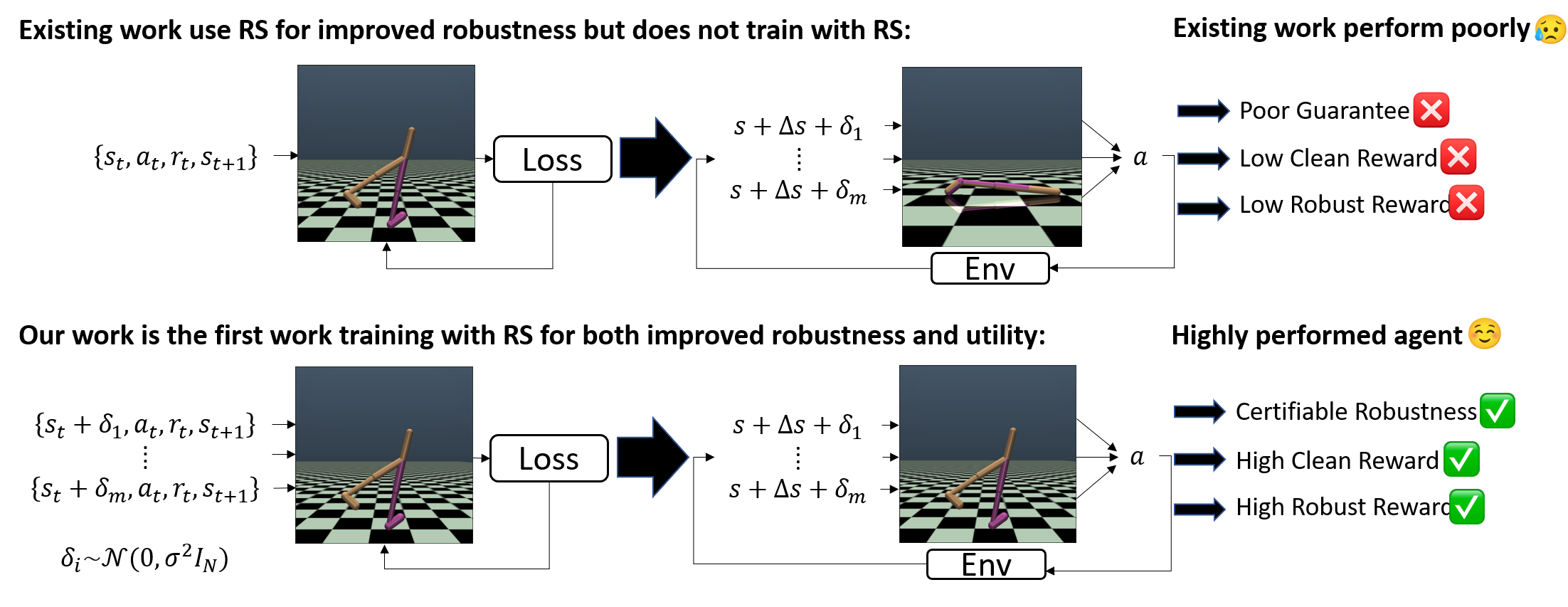}
\vspace{-20pt}
\caption{The overview of our framework. We propose new DRL training algorithms leveraging Randomized Smoothing, achieving strong certifiable robustness, high clean reward, and high robust reward simultaneously. }
\label{fig:overview}
\end{figure*}
\section{Learning Robust DRL Agents with Randomized Smoothing}
\label{sec:our approach}

In this section, we propose first training algorithms leveraging Randomized Smoothing (RS) to achieve certifiably robust agents, solving the issues mentioned in Section \ref{sec:trade off} and effectively boosting the robustness as shown in Figure \ref{fig:trade-off}. The overview of our framework is shown in Figure \ref{fig:overview}. Our primary focus centers on two representative RL algorithms: DQN for discrete action space, and PPO for continuous action space, which are the focus of prior works in robust DRL literature \cite{sa,radialrl,wocar, optimalattack,paad,crop, policysmoothing}. 

\subsection{S-DQN (Smoothed - Deep Q Network)}
\label{sec:sdqn}

% In the supervised learning setting, a large smoothing variance is introduced to improve the robustness of the smoothed classifiers. However, based on the experiments of Figure \ref{fig:trade-off}, we found that none of the smoothed DQN agents can tolerate the large noise introduced by RS and perform even worse than the non-smoothed robust agents. Hence, it is essential to develop a strategy to mitigate the effect of the noise from RS, which motivates us to leverage the Denoised Smoothing strategy.
We describe the details of training, testing, and evaluating S-DQN in the following paragraphs. 

\begin{figure*}[!t]
\centering
\includegraphics[width=1\textwidth]{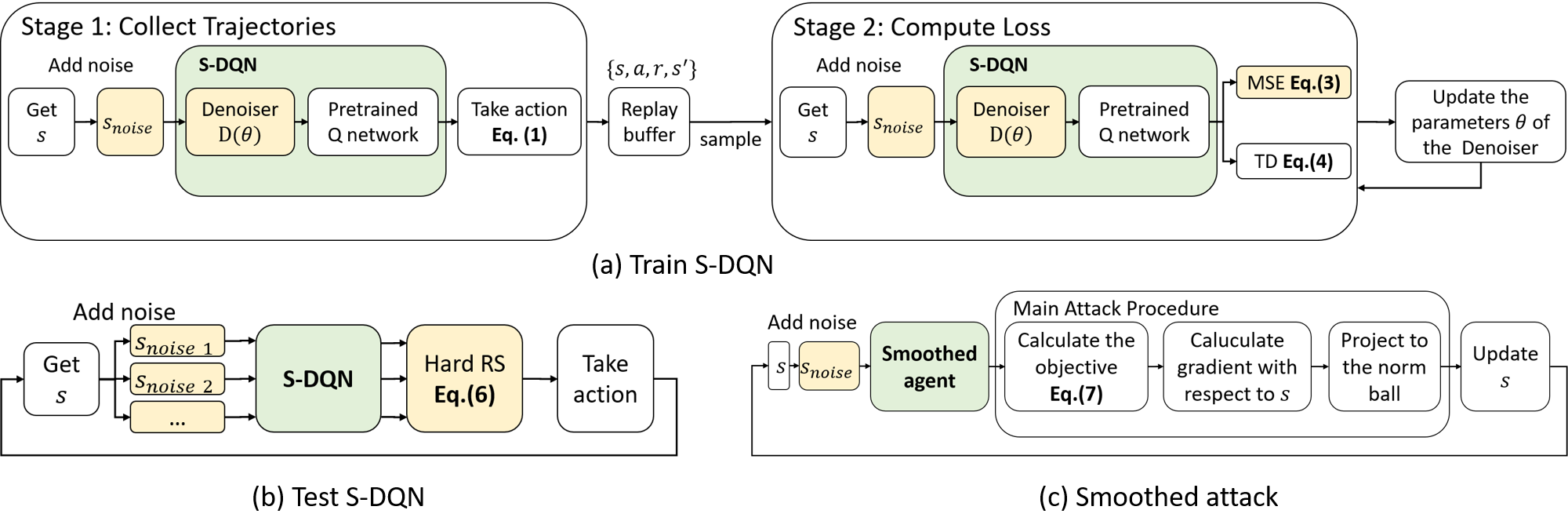}
\vspace{-20pt}
\caption{The flow chart of: (a) training process of S-DQN, (b) testing process of S-DQN, (c) our Smoothed Attack pipeline for smoothed agents, which is much more effective than non-smoothed attack.}
\label{fig:sdqn}
\end{figure*}

\paragraph{Training and loss function.}
The training process of S-DQN is shown in Figure \ref{fig:sdqn} (a), which involves two main steps: collecting transitions and updating the networks. First, we collect the transitions $\{s_t, a_t, r_t, s_{t+1}\}$ with noisy states, which can be formulated as follows:
\begin{equation}
    a_t=
    \begin{cases}
        \argmax_a Q(D(\tilde{s}_t;\theta),a),\;\textrm{with probability}\;1-\epsilon \\
        \textrm{Random Action},\;\textrm{with probability}\;\epsilon
    \end{cases}
\end{equation}
where $\tilde{s}_t$ is the state with noise $\tilde{s}_t=s_t+\mathcal{N}(0,\sigma^2I_N)$, $D$ is the denoiser, $Q$ is the pretrained Q-network, and $\sigma$ is the standard deviation of the Gaussian distribution. Here, we introduce a denoiser $D$ before the Q-network, aiming to alleviate the side effects of the low clean reward resulting from the noisy states. After collecting the transitions, they are stored in the replay buffer. In the second stage, we sample some transitions from the replay buffer and update the parameters of the denoiser $D$. The entire loss function is designed with two parts, reconstruction loss $\mathcal{L}_\textrm{R}$ and temporal difference loss $\mathcal{L}_{\textrm{TD}}$:
\begin{equation}
\label{e:7}
    \mathcal{L}=\lambda_1\mathcal{L}_\textrm{R}+\lambda_2\mathcal{L}_{\textrm{TD}},
\end{equation}
where $\lambda_1$ and $\lambda_2$ are the hyperparameters. Suppose the sampled transition is $\{s,a,r,s^\prime\}$, the reconstruction loss $\mathcal{L}_\textrm{R}$ is defined as:
\begin{equation}
    \mathcal{L}_\textrm{R} =\frac{1}{N}||D(\tilde{s};\theta)-s||^2_2,
\end{equation}
where $\tilde{s}=s+\mathcal{N}(0,\sigma^2I_N)$, and $N$ is the dimension of the state. The reconstruction loss is the mean square error (MSE) between the original state and the output of the denoiser. This loss aims to train the denoiser $D$ to effectively reconstruct the original state. The temporal difference loss $\mathcal{L}_{\textrm{TD}}$ is defined as:
\begin{equation}
    \begin{split}
        & \mathcal{L}_{\textrm{TD}} = 
        \begin{cases}
            \frac{1}{2\zeta}\eta^2,\;\textrm{if}\;|\eta|<\zeta \\
            |\eta|-\frac{\zeta}{2},\;\textrm{otherwise}
        \end{cases} \\
        & \eta=r+\gamma\max_{a^\prime}Q(s^\prime,a^\prime)-Q(D(\tilde{s};\theta),a),
    \end{split}
\end{equation}
where $\zeta$ is set to $1$. Our designed $\mathcal{L}_{\textrm{TD}}$ is different from the common temporal difference loss in the DQN learning: the current Q-value is estimated with the denoised state (the output of $D$) and the target Q-value remains clean without noisy input. Note that the pretrained Q-network $Q$ can be replaced with robust agents such as RadialDQN \cite{radialrl} and our S-DQN framework can also be combined with adversarial training to further improve the robustness. We will discuss this later in Section \ref{sec:experiment}. The full training algorithm can be found in Appendix \ref{sec:Training Algorithm of SDQN} Algorithm \ref{alg:training SDQN}.

\paragraph{Testing with hard randomized smoothing.} The testing process of S-DQN is shown Figure \ref{fig:sdqn} (b). In the testing stage, we need to obtain the smoothed Q-values of S-DQN. We leverage the hard Randomized Smoothing (hard RS) strategy to enhance robustness, which will be further discussed in Section \ref{sec:certification}. We first define the hard Q-value $Q_h$ as follows:
\begin{equation}
    \label{e:hard}
    Q_h(s,a)=\mathds{1}_{\{a=\argmax_{a^\prime}Q(s,a^\prime)\}}
\end{equation}
Note that the hard Q-value $Q_h$ is always in $[0,1]$. Then, we define the hard RS for S-DQN as follows:
\begin{equation}
\label{e:11}
    \widetilde{Q}(s,a)=\mathbb{E}_{\delta\sim\mathcal{N}(0,\sigma^2I_N)}Q_h(D(s+\delta),a).
\end{equation}
In practice, we need to estimate the expectation to get $\widetilde{Q}$, which can be done by using Monte Carlo sampling. The action is then selected by taking  $\argmax_{a}\widetilde{Q}(s,a)$. The full algorithm is in Appendix \ref{sec:Testing Algorithm of SDQN} Algorithm \ref{alg:test SDQN}.

\paragraph{New attack framework: Smoothed attack.} In \cite{crop}, they evaluated all the smoothed DQN agents with the classic Projected Gradient Descent (PGD) attack. However, we found that the classic PGD attack is ineffective in decreasing the reward of the smoothed DQN agents as shown in Table \ref{table:new attack}. Hence, we propose a new attack framework named Smoothed Attack, which is specifically designed for the smoothed agents to evaluate our S-DQN. The pipeline of Smoothed Attack is shown in Figure \ref{fig:sdqn} (c). The objective of Smoothed Attack is as follows:
\begin{equation}
\label{e:attack}
    \begin{split}
        \min_{\Delta s}\log\dfrac{\exp{Q(D(\tilde{s}+\Delta s),a^*)}}{\Sigma_a\exp{Q(D(\tilde{s}+\Delta s),a)}},\;\textrm{s.t.}\;||\Delta s||_{p}\leq\epsilon,
    \end{split}
\end{equation}
where $a^*=\argmax_{a}\widetilde{Q}(s,a)$, $\widetilde{Q}(s,a)$ is defined in Eq.(\ref{e:11}), $\tilde{s}=s+\mathcal{N}(0,\sigma^2I_N)$, $\epsilon$ is the attack budget, and $p=2\;\textrm{or}\;\infty$ in our setting. In our Smoothed Attack, the state with perturbation is added with a noise sampled from Gaussian distribution with the corresponding smoothing variance $\sigma$. This setting can be integrated with various existing attacks, such as PGD attack and PA-AD \cite{paad}, by replacing the objective with the Smoothed Attack objective in Eq.(\ref{e:attack}). The comparison of our Smoothed Attack (S-PGD and S-PA-AD) against the PGD attack and PA-AD attack is in Table \ref{table:new attack}. The full algorithm of our smoothed attack is in Appendix \ref{sec:Attack Algorithm of SDQN} Algorithm \ref{alg:PGD attack designed for SDQN}.

\begin{table}[!t]
\vspace{-5pt}
\caption {The comparison between our smoothed attacks (S-PGD and S-PA-AD) and the existing attacks. A lower reward means the attack is stronger. Our S-PGD attack reduces $61.8\%$ of the reward of S-DQN on average, which is over $2.62\times$ stronger than $23.6\%$ of the classic PGD attack. Our S-PA-AD attack reduces $55.4\%$ of the reward of S-DQN on average, which is over $1.15\times$ stronger than $48.1\%$ of the original version of PA-AD attack. The $\ell_{\infty}$ budget is set to $\epsilon=0.05$ in all the attacks.}
\label{table:new attack}
\centering
\tabcolsep=0.05cm
\tiny
\begin{tabular*}{\linewidth} {@{\extracolsep{\fill}} l|llll}
\toprule
    \multicolumn{1}{l}
    {Agents} & Environments & No Attack & classic PGD Attack & \textbf{S-PGD Attack (Ours)} \\
    \midrule
    \multicolumn{1}{l} {S-DQN} & Pong & $20.4 \pm 0.5$ & $19.4 \pm 2.1$ & $\bf{18.4 \pm 2.1}$\\
    \multicolumn{1}{l} {} & Freeway & $34.0 \pm 0.0$ & $32.0 \pm 1.4$ & $\bf{6.6 \pm 2.2}$ \\
    \multicolumn{1}{l} {} & RoadRunner & $47480 \pm 8807$ & $ 17740 \pm 3718$ & $\bf{0 \pm 0}$ \\
\midrule
    \multicolumn{1}{l}
    {Agents} & Environments & No Attack & PA-AD & \textbf{S-PA-AD (Ours)} \\
    \midrule
    \multicolumn{1}{l} {S-DQN} & Pong & $20.4 \pm 0.5$ & $19.4 \pm 0.8$ & $\bf{18.6 \pm 1.2}$ \\
    \multicolumn{1}{l} {} & Freeway & $34.0 \pm 0.0$ & $19.8 \pm 1.5$ & $\bf{13.0 \pm 2.1}$ \\
    \multicolumn{1}{l} {} & RoadRunner & $47480 \pm 8807$ & $\bf{0 \pm 0}$ & $\bf{0 \pm 0}$ \\
\bottomrule
\end{tabular*}
\end{table}

\subsection{S-PPO (Smoothed - Proximal Policy Optimization)}
\label{sec:sppo}

\begin{figure*}[!t]
\centering
\includegraphics[width=1\textwidth]{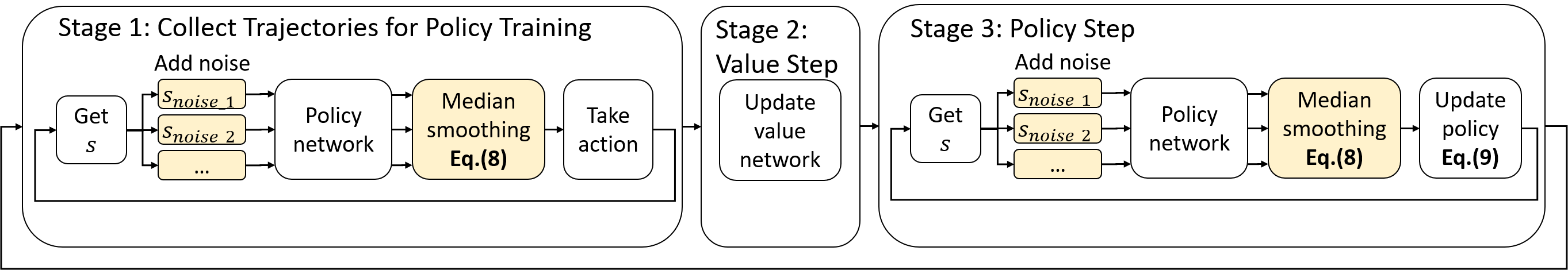}
\vspace{-20pt}
\caption{The training process of S-PPO.}
\label{fig:train sppo}
\end{figure*}

The specifics of training, testing, and evaluating our proposed S-PPO are outlined in the following paragraphs.

\paragraph{Training and loss function.} PPO agents demonstrate enhanced tolerance to Gaussian noise in contrast to DQN agents. This attribute allows us to directly employ RS for training the PPO agents. The training process of S-PPO is shown in Figure \ref{fig:train sppo}. Initially, we gather trajectories using the smoothed policy and subsequently update both the value network and the policy network. In the trajectory collection phase, We use the Median Smoothing \cite{median} strategy to smooth our agents. The median value has a nice property: it is almost unaffected by the outliers. Hence, Median Smoothing can give a better estimation of the expectation than mean smoothing when the number of samples is small. The smoothed policy of S-PPO is defined as follows:
\begin{equation}
\label{e:13}
    \tilde{\pi}_i(a|s)=\mathcal{N}(\widetilde{M}_i,\widetilde{\Sigma}_i^2),\;\forall i\in\{1,...,N_{\textrm{action}}\}
\end{equation}
where $\widetilde{M}_i=\textrm{sup}\{M\in\mathbb{R}|\mathbb{P}_{\delta\sim\mathcal{N}(0,\sigma^2I_N)}[a^{\textrm{mean}}_i\leq M]\leq p\}$, $\widetilde{\Sigma}_i=\textrm{sup}\{\Sigma\in\mathbb{R}|\mathbb{P}_{\delta\sim\mathcal{N}(0,\sigma^2I_N)}[a^{\textrm{std}}_i\leq \Sigma]\leq p\}$, $(a^{\textrm{mean}}_i,a^{\textrm{std}}_i)$ is the output of policy network given a state with noise $s+\delta$ as input, which represents the mean and standard deviation of the $i$-th coordinate of the action, $N_{\textrm{action}}$ is the dimension of the action, and $p$ is the percentile. 

Now, we define the loss function for S-PPO as follows: 
\begin{equation}
   \begin{aligned}
       & \mathcal{L}_{\tilde{\pi}}(\theta)=-\mathbb{E}_t[\min(\mathcal{R}_{\tilde{\pi}}\hat{A}_t,\textrm{clip}(\mathcal{R}_{\tilde{\pi}},1-\epsilon_{\textrm{c}},1+\epsilon_{\textrm{c}})\hat{A}_t)], \\
       & \mathcal{R}_{\tilde{\pi}}=\frac{\tilde{\pi}(a_t|s_t;\theta)}{\tilde{\pi}(a_t|s_t;\theta_{\textrm{old}})},
   \end{aligned}
\end{equation}
where $\hat{A}_t$ is the advantage, and $\epsilon_{\textrm{c}}$ is the clipping hyperparameter. This is the loss of the classic PPO algorithm combined with RS. Note that our S-PPO can also be combined with robust PPO algorithms such as SGLD \cite{sa}, Radial \cite{radialrl}, or WocaR \cite{wocar}.

\paragraph{Adversary training for S-PPO.} In ATLA-PPO \cite{optimalattack} and PA-ATLA-PPO \cite{paad}, they jointly train a policy network and an adversarial network to robustify the PPO agents. Our S-PPO can also be combined with these adversarial training methods by modifying the adversarial policy and objective to align with the smoothed one. The smoothed adversarial policy is defined as follows:
\begin{equation}
    \tilde{\mathcal{A}}_i(\Delta p|s)=\mathcal{N}(\widetilde{M}_i,\widetilde{\Sigma}_i^2),\;\forall i\in\{1,...,N_{\Delta p}\}
\end{equation}
where $\mathcal{A}$ is the adversary, $\Delta p$ is the attack direction, $\widetilde{M}_i=\textrm{sup}\{M\in\mathbb{R}|\mathbb{P}_{\delta\sim\mathcal{N}(0,\sigma^2I_N)}[\Delta p^{\textrm{mean}}_i\leq M]\leq p\}$, $\widetilde{\Sigma}_i=\textrm{sup}\{\Sigma\in\mathbb{R}|\mathbb{P}_{\delta\sim\mathcal{N}(0,\sigma^2I_N)}[\Delta p^{\textrm{std}}_i\leq \Sigma]\leq p\}$, $(\Delta p^{\textrm{mean}}_i,\Delta p^{\textrm{std}}_i)$ is the output of the adversarial network given a state with noise $s+\delta$ as input, which represents the mean and standard deviation of the $i$-th coordinate of the perturbation, and $N_{\Delta p}$ is the dimension of $\Delta p$.

The loss of training smoothed adversarial policy is defined as follows, which is designed to minimize the surrogate reward:
\begin{equation}
   \begin{aligned}
        & \mathcal{L}_{\tilde{\mathcal{A}}}(\theta)=\mathbb{E}_t[\min(\mathcal{R}_{\tilde{\mathcal{A}}}\hat{A}_t,\textrm{clip}(\mathcal{R}_{\tilde{\mathcal{A}}},1-\epsilon_{\textrm{c}},1+\epsilon_{\textrm{c}})\hat{A}_t)], \\
        & \mathcal{R}_{\tilde{\mathcal{A}}}=\frac{\tilde{\mathcal{A}}(\Delta p_t|s_t;\theta)}{\tilde{\mathcal{A}}(\Delta p_t|s_t;\theta_{\textrm{old}})}.
   \end{aligned}
\end{equation}
In ATLA, $\Delta p$ represents the direction of the state change $\Delta s$ used to perturb the state of the PPO agents. On the other hand, in PA-ATLA, $\Delta p$ represents the direction of the action change $\Delta a$. To induce the PPO agents to undergo the specified action change $\Delta a$, a Fast Gradient Sign Method (FGSM) attack is then executed to perturb the state. We use S-FGSM, which is the Smoothed Attack, while using the PA-ATLA algorithm to perform adversarial training for S-PPO.

The full algorithm of training S-PPO is in Appendix \ref{sec:Training Algorithm of SPPO} Algorithm \ref{alg:training SPPO} and \ref{alg:collect trajectories}.

\paragraph{Testing.}
We also use Median Smoothing during testing to obtain the smoothed policy. However, we use the smoothed deterministic policy as follows:
\begin{equation}
    \tilde{\pi}_{i,\textrm{det}}(s)=\widetilde{M}_i,\;\forall i\in\{1,...,N_{\textrm{action}}\},
\end{equation}
where $\widetilde{M}_i=\textrm{sup}\{M\in\mathbb{R}|\mathbb{P}_{\delta\sim\mathcal{N}(0,\sigma^2I_N)}[a^{\textrm{mean}}_i\leq M]\leq p\}$, and $a^{\textrm{mean}}_i$ is the output of policy network given a state with noise $s+\delta$ as input ($a^{\textrm{mean}}_i=\pi_{i,\textrm{det}}(s+\delta)$) representing the mean of the $i$-th coordinate of the action. Here we only use the $a^{\textrm{mean}}$ value of the output of the policy network for smoothing.

\paragraph{Attack.}
To evaluate the performance of our S-PPO, we use the Maximal Action Difference (MAD) Attack and Minimum Robust Sarsa (Min-RS) Attack proposed in \citet{sa}. Furthermore, We also evaluate our S-PPO under the two strongest optimal adversaries \cite{optimalattack,paad}. \cite{optimalattack} proposed the Optimal Attack, employing an adversarial agent to perturb the states. \cite{paad} proposed the state-of-the-art PA-AD attack, where an adversarial agent determines a direction and uses FGSM to perturb the states based on the specified direction. In the PPO setting, we did not find a significant difference between the smoothed attack and the non-smoothed attack (see Table \ref{table:smoothed attack ppo} in Appendix \ref{sec: additional experiments}), and hence, we used the original setting for every attack.
\section{Robustness certification}
\label{sec:certification}
The strength of the smoothed agents lies in their certifiable robustness. However, previous literature \cite{crop,policysmoothing} fails to give a good expression for the certified radius of DQN agents and has not derived the action bound for PPO agents. To make the study of certifiable robustness more complete, we formally formulate the certified radius, action bound, and reward lower bound of our S-DQN and S-PPO agents.

\paragraph{Certified Radius for S-DQN.}
The certified radius for our S-DQN is defined as follows:
\begin{equation}
\label{e:CR}
    R_t=\dfrac{\sigma}{2}(\Phi^{-1}(\widetilde{Q}(s_t,a_1))-\Phi^{-1}(\widetilde{Q}(s_t,a_2))),
\end{equation}
where $a_1$ is the action with the largest Q-value among all the other actions, $a_2$ is the ”runner-up” action, $R_t$ is the certified radius at time $t$, $\Phi$ is the CDF of normal distribution, $\sigma$ is the smoothing variance, and $\widetilde{Q}(s,a)$ is defined in Eq.(\ref{e:11}). As long as the $\ell_2$ perturbation is bounded by $R_t$, the action is guaranteed to be the same. 

Note that our expression of the certified radius is different from the one proposed in CROP \cite{crop} since we use hard RS. In CROP, they took the average of the output samples, which is the mean smoothing strategy. However, this might not lead to a precise estimation of the certified radius since it requires estimating the output range $[V_{\textrm{min}},V_{\textrm{max}}]$ of the Q-network. The certified radius proposed in CROP is shown as follows:
\begin{equation}
\label{e:crop}
    \begin{aligned}
        R_t= & \dfrac{\sigma}{2}(\Phi^{-1}(\dfrac{\widetilde{Q}_{\textrm{CROP}}(s_t,a_1)-\Delta-V_{\textrm{min}}}{V_{\textrm{max}}-V_{\textrm{min}}}) \\
        & -\Phi^{-1}(\dfrac{\widetilde{Q}_{\textrm{CROP}}(s_t,a_2)+\Delta-V_{\textrm{min}}}{V_{\textrm{max}}-V_{\textrm{min}}})),
    \end{aligned}
\end{equation}
where $R_t$ is the certified radius at time step $t$, $Q_{\textrm{CROP}}:\mathcal{S}\times\mathcal{A}\rightarrow[V_{\textrm{min}},V_{\textrm{max}}]$, $\widetilde{Q}_{\textrm{CROP}}(s,a)=\frac{1}{m}\Sigma_{i=1}^mQ_{\textrm{CROP}}(s+\delta_i,a)$, $\delta_i\sim\mathcal{N}(0,\sigma^2I_N),\forall i\in\{1,...,m\}$, $a_1$ is the action with the largest Q-value, $a_2$ is the "runner-up" action, $\Delta=(V_{\textrm{max}}-V_{\textrm{min}})\sqrt{\frac{1}{2m}\ln\frac{1}{\alpha}}$, $\Phi$ is the CDF of standard normal distribution, $m$ is the number of the samples, and $\alpha$ is the one-side confidence parameter. Based on this expression, the output range of the Q-network $[V_{\textrm{min}},V_{\textrm{max}}]$ can significantly affect the certified radius. The certified radius is small when the output range of the Q-network $[V_{\textrm{min}},V_{\textrm{max}}]$ is large (e.g. Suppose $\widetilde{Q}_{\textrm{CROP}}(s_t,a_1)=3$, $\widetilde{Q}_{\textrm{CROP}}(s_t,a_2)=-3$, $\sigma=0.1$, $m=100$, and $\alpha=0.05$. The certified radius is only $0.007$ under $[V_{\textrm{min}},V_{\textrm{max}}]=[-10,10]$. Instead, if we narrow down the interval to $[V_{\textrm{min}},V_{\textrm{max}}]=[-3.5,3.5]$, the certified radius grows to $0.086$). CROP estimated $[V_{\textrm{min}},V_{\textrm{max}}]$ by sampling some trajectories and finding the maximum and the minimum of the Q-values. However, if the actual interval is much larger than the estimation (which is likely to happen in practice since it is impossible to go over all the states), the certified radius can be significantly overestimated.

In contrast, our hard RS strategy eliminates the need for estimating $[V_{\textrm{min}},V_{\textrm{max}}]$, resulting in a more precise estimation of the certified radius. Moreover, based on Eq.(\ref{e:CR}), the certified radius of our S-DQN is not influenced by the out range of the Q-network $[V_{\textrm{min}},V_{\textrm{max}}]$, which gives a more stable guarantee. Detailed experiments for the certified radius of our S-DQNs versus the CROP agents are provided in Appendix \ref{sec:cr result}, demonstrating that our S-DQNs achieve a larger radius. The proof of the certified radius for S-DQN can be found in Appendix \ref{sec:CR proof DQN}.

\vspace{-2pt}
\paragraph{Action Bound for S-PPO.}
Unfortunately, unlike the discrete action setting, there is no guarantee that the action will not change under a certain radius in the continuous action setting. Hence, we derive the \textbf{Action Bound}, which bounds the policy of S-PPO agents in a close region:
\begin{equation}
    \tilde{\pi}_{\textrm{det},{\underline{p}}}(s_t) \preceq \tilde{\pi}_{\textrm{det},p}(s_t+\Delta s)\preceq\tilde{\pi}_{\textrm{det},{\overline{p}}}(s_t),\;\textrm{s.t.}\;||\Delta s||_2\leq\epsilon,
\end{equation}
where $\tilde{\pi}_{i,\textrm{det},p}(s)=\textrm{sup}\{a_i\in\mathbb{R}|\mathbb{P}_{\delta\sim\mathcal{N}(0,\sigma^2I_N)}[\pi_{i,\textrm{det}}(s+\delta)\leq a_i]\leq p\},\forall i\in\{1,...,N_{\textrm{action}}\}$, $\underline{p}=\Phi(\Phi^{-1}(p)-\frac{\epsilon}{\sigma})$, $\overline{p}=\Phi(\Phi^{-1}(p)+\frac{\epsilon}{\sigma})$, and $p$ is the percentile. We designed a metric based on this action bound to evaluate the certified robustness of S-PPO agents. See Appendix \ref{sec:adiv result} for more details. The proof of the action bound can be found in Appendix \ref{sec:AB proof PPO}.

\vspace{-2pt}
\paragraph{Reward lower bound for smoothed agents.}
% By viewing the whole trajectory as a function $F_\pi$, we define $F_\pi :\mathbb{R}^{H\times N}\rightarrow\mathbb{R}$ as follows:
% \begin{equation}
%     F_\pi(\boldsymbol{\Delta s})=\Sigma^{H-1}_{t=0}R(s_t,\pi(s_t+\Delta s_t)),\;\textrm{s.t.}\;s_{t+1}\sim P(s_t,\pi(s_t+\Delta s_t))
% \end{equation}
% where $\boldsymbol{\Delta s}=[\Delta s_0,...,\Delta s_{H-1}]^T$, $R$ is the reward function, $P$ is the transition, and $H$ is the length of the trajectory. Note that $F_\pi(\boldsymbol{0})$ represents the clean reward of the whole trajectory. Again, we use median smoothing to obtain the smoothed version of $F_\pi$:
% \begin{equation}
%     \widetilde{F}_{\pi,p}(\boldsymbol{0})=\textrm{sup}\{r\in\mathbb{R}|\mathbb{P}_{\boldsymbol{\delta}\sim\mathcal{N}(0,\sigma^2I_{H\times N})}[F_\pi(\boldsymbol{\delta})\leq r]\leq p\}
% \end{equation}
% where $\boldsymbol{\delta}=[\delta_0,...,\delta_{H-1}]^T$, $\delta_i\sim\mathcal{N}(0,\sigma^2I_N),\forall i\in\{0,...,H-1\}$, and $p$ is the percentile set to $0.5$. 
By viewing the whole trajectory as a function $F_\pi$, we define $F_\pi :\mathbb{R}^{H\times N}\rightarrow\mathbb{R}$ that maps the vector of perturbations for the whole trajectory $\boldsymbol{\Delta s}=[\Delta s_0,...,\Delta s_{H-1}]^\top$ to the cumulative reward. Then, the reward lower bound is defined as follows:
\begin{equation}
    \widetilde{F}_{\pi,p}(\boldsymbol{\Delta s})\geq\widetilde{F}_{\pi,\underline{p}}(\boldsymbol{0}),\;\textrm{s.t.}\;||\boldsymbol{\Delta s}||_2\leq B,
\end{equation}
where $\widetilde{F}_{\pi,p}(\boldsymbol{\Delta s})=\textrm{sup}\{r\in\mathbb{R}|\mathbb{P}_{\boldsymbol{\delta}\sim\mathcal{N}(0,\sigma^2I_{H\times N})}[F_\pi(\boldsymbol{\delta}+\boldsymbol{\Delta s})\leq r]\leq p\}$, $\widetilde{F}_{\pi,\underline{p}}(\boldsymbol{0})=\textrm{sup}\{r\in\mathbb{R}|\mathbb{P}_{\boldsymbol{\delta}\sim\mathcal{N}(0,\sigma^2I_{H\times N})}[F_\pi(\boldsymbol{\delta})\leq r]\leq \underline{p}\}$, $\boldsymbol{\delta}=[\delta_0,...,\delta_{H-1}]^\top$, $\underline{p}=\Phi(\Phi^{-1}(p)-\frac{B}{\sigma})$, $H$ is the length of the trajectory, and $B$ is the $\ell_2$ attack budget for the entire trajectory. If the attack budget of each state is $\epsilon$, then $B=\epsilon\sqrt{H}$. This bound ensures that the reward will not fall below a certain value while given any $\ell_2$ perturbation with budget $B$. We will discuss the reward lower bound for all the smoothed agents in Section \ref{sec:experiment}. The proof of the reward lower bound can be found in Appendix \ref{sec:reward lower bound proof}. 

In practice, it is necessary to introduce the confidence interval, which can change the bounds based on the sample number, while estimating all the bounds introduced above. The details of estimating the bounds are provided in Appendix \ref{sec:detail of estimating bounds}.
\section{Experiment}
\label{sec:experiment}

\paragraph{Setup.}
We follow the previous robust DRL literature to conduct experiments on Atari \cite{atari} and Mujoco \cite{gym} benchmarks. In our DQN settings, the evaluations are done in three Atari environments — Pong, Freeway, and RoadRunner. We train the denoiser $D$ with different base agents and with adversarial training. Our methods are listed as follows:
\begin{itemize}
    \item S-DQN (\emph{\{Base agent\}}): S-DQN combined with a certain base agents. \emph{\{base agent\}} can be Radial \cite{radialrl} or Vanilla (simple DQN).
    \vspace{-5pt}
    \item S-DQN (S-PGD): S-DQN (Vanilla) adversarially trained with our proposed S-PGD.
\end{itemize}
We compare our S-DQN with the following baselines:
\begin{itemize}
    \item Non-smoothed robust agents: WocaRDQN \cite{wocar}, RadialDQN \cite{radialrl}, SADQN \cite{sa}.
    \vspace{-5pt}
    \item Previous smoothed agents \cite{crop, policysmoothing}: WocaRDQN+RS, RadialDQN+RS, SADQN+RS. We use \emph{\{base agent\}}+RS to denote them.
\end{itemize}
In our PPO settings, the evaluations are done on two continuous control tasks in the Mujoco environments — Walker and Hopper. We train each agent $15$ times and report the median performance as suggested in \citet{sa} since the training variance of PPO algorithms is high. Our methods are listed as follows:
\begin{itemize}
    \item S-PPO (\emph{\{base algorithm\}}): S-PPO combined with a certain base algorithms. \emph{\{base algorithm\}} can be SGLD \cite{sa}, Radial \cite{radialrl}, WocaR \cite{wocar}, or Vanilla (simple PPO).
    \vspace{-5pt}
    \item S-PPO (S-ATLA), S-PPO (S-PA-ATLA): S-PPO with smoothed adversarial training described in Section \ref{sec:sppo} "\textit{Adversary training for S-PPO}".
\end{itemize}
We compare our S-PPO with the following baselines:
\begin{itemize}
    \item Non-smoothed robust agents: WocaRPPO \cite{wocar}, PA-ATLAPPO \cite{paad}, ATLAPPO \cite{optimalattack}, RadialPPO \cite{radialrl}, SGLDPPO \cite{sa}.
    \vspace{-5pt}
    \item Previous smoothed agents: WocaRPPO+RS, PA-ATLAPPO+RS, ATLAPPO+RS, RadialPPO+RS, SGLDPPO+RS.
\end{itemize}
See Appendix \ref{sec:detailed setting for DQN and PPO} for more details about our setting.

\paragraph{Robust reward and lower bound for S-DQN.} The robust reward of our S-DQN under $\ell_{\infty}$ PGD attack and PA-AD attack \cite{paad} is shown in Table \ref{table:dqn reward}. The presented rewards are first normalized and then averaged across the three environments. Note that we use our stronger S-PGD and S-PA-AD introduced in Section \ref{sec:sdqn} to evaluate all the smoothed agents. Our S-DQN (Radial), S-DQN (S-PGD), and S-DQN (Vanilla) exhibit superior performance compared to the state-of-the-art robust RadialDQN and WocaRDQN. Notably, our S-DQN (Vanilla) already demonstrates greater robustness than RadialDQN without further combining with other robust agents. The poor performance of rows (c) suggests that the previous smoothed agents struggle to tolerate the Gaussian noise introduced by RS and fail to enhance the reward under attack. More detailed experiment results and discussion about the robust reward for S-DQN can be found in Appendix \ref{sec: detailed robust reward for S-DQN}.

\begin{figure}[t]
\centering
\vspace{7pt}
\includegraphics[width=0.4\textwidth]
{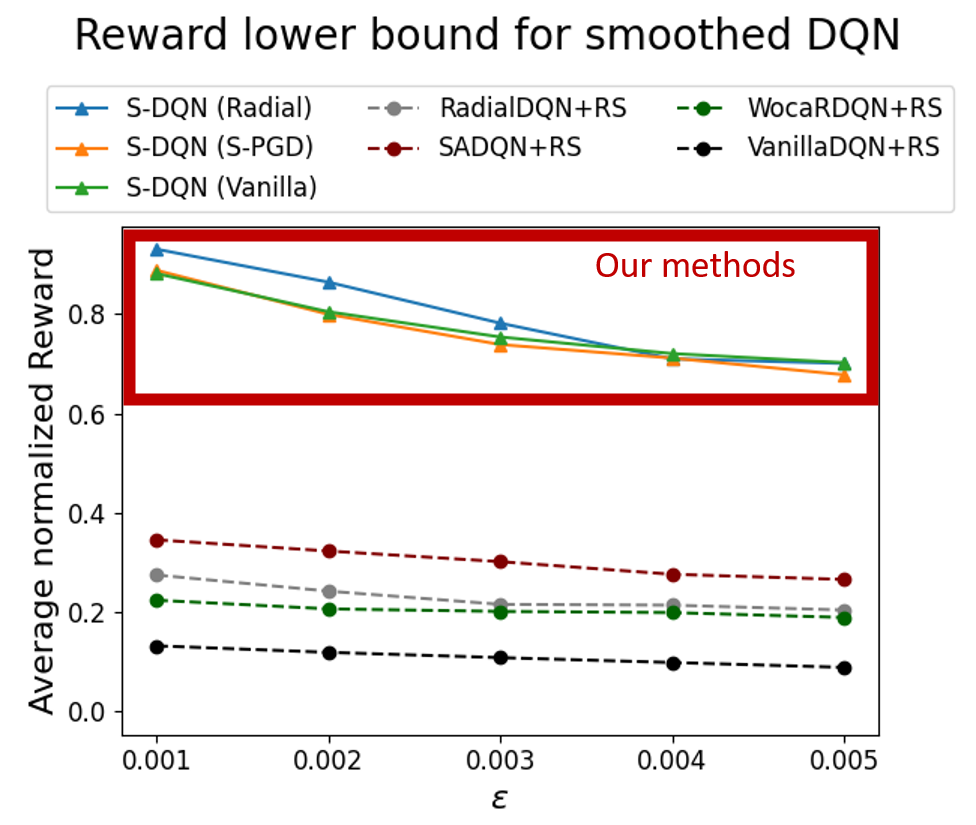}
\vspace{-10pt}
\caption{The certified reward lower bound of smoothed DQN agents. Our S-DQNs achieve a much higher lower bound than all the previous smoothed agents.}
\label{fig:lower bound DQN}
\end{figure}

Figure \ref{fig:lower bound DQN} shows the reward lower bound of our S-DQNs. Our S-DQNs exhibit high reward lower bounds compared to the previous smoothed agents, indicating that our method can enhance not only the empirical robustness but also the robustness guarantee. More detailed experiment results for the reward lower bound can be found in Appendix \ref{sec: detailed reward lower bound for DQN}.

\paragraph{Robust reward and lower bound for S-PPO.} The robust reward of our S-PPO under attacks is shown in Table \ref{table:ppo reward}. The presented rewards are first normalized and then averaged across the two environments. Our S-PPO agents constantly outperform their counterparts (previous smoothed agents and the SOTA robust agents) for all robust training algorithms. Through comparing rows (b) and (c), the previous smoothed agents exhibit lower clean reward and only marginal improvement on the reward under attacks, suggesting that naively applying RS during the test time cannot improve the robustness of PPO agents. In addition, our S-PPO agents receive a much higher clean reward on average, showing that our RS training approach can further boost performance in the non-adversarial setting. More detailed experiment results and discussion about the robust reward for S-PPO can be found in Appendix \ref{sec: detailed robust reward for S-PPO}.

Our S-PPOs also exhibit higher reward lower bounds than the previous smoothed PPO agents, which is shown in Figure \ref{fig:lower bound PPO}. More detailed experiment results for the reward lower bound can be found in Appendix \ref{sec: detailed reward lower bound for PPO}.

\begin{table}[!t]
\caption {The average normalized reward of DQN agents under  $\ell_{\infty}$ PGD attack and PA-AD attack. Our S-DQNs achieve the highest robust reward, especially under a large attack budget $\epsilon$.}
\label{table:dqn reward}
\tabcolsep=0.015cm
\centering
\tiny
\begin{NiceTabular*}{\linewidth}{@{\extracolsep{\fill}} lccccccc}[colortbl-like]
\toprule
    {Avg normalized reward} 
    & \multicolumn{1}{c} {Clean} & \multicolumn{5}{c} {PGD attack} & \multicolumn{1}{c} {PA-AD attack}\\
    \cmidrule(lr){3-7}
    {$\epsilon(\ell_{\infty})$} & & $0.01$ & $0.02$ & $0.03$ & $0.04$ & $0.05$ & $0.05$ \\
    \midrule
    \rowcolor{lightblue}{\textbf{(a) Ours:}} \\
    \rowcolor{lightblue}{S-DQN (Radial)} & $0.929$ & $0.928$ & $\bf{0.932}$ & $\bf{0.830}$ & $\bf{0.788}$ & $\bf{0.735}$ & $\bf{0.669}$ \\
    \rowcolor{lightblue}{S-DQN (S-PGD)} & $0.945$ & $0.945$ & $0.886$ & $0.775$ & $0.700$ & $0.450$ & $0.552$ \\
    \rowcolor{lightblue}{S-DQN (Vanilla)} & $0.989$ & $0.818$ & $0.660$ & $0.601$ & $0.498$ & $0.377$ & $0.442$ \\
    \midrule
    \multicolumn{2}{l} {\textbf{(b) SOTA robust agents:}} \\
    {RadialDQN} & $0.926$ & $\bf{0.947}$ & $0.770$ & $0.337$ & $0.206$ & $0.210$ & $0.248$ \\
    {SADQN} & $0.949$ & $0.825$ & $0.302$ & $0.205$ & $0.207$ & $0.185$ & $0.224$ \\
    {WocaRDQN} & $0.865$ & $0.617$ & $0.218$ & $0.204$ & $0.208$ & $0.216$ & $0.210$ \\
    {VanillaDQN} & $\bf{1.000}$ & $0.000$ & $0.000$ & $0.000$ & $0.000$ & $0.000$ & $0.000$ \\
    \multicolumn{4}{l} {\textbf{(c) Previous smoothed agents:}} \\
    {RadialDQN+RS} & $0.310$ & $0.295$ & $0.281$ & $0.264$ & $0.271$ & $0.240$ & $0.265$ \\
    {SADQN+RS} & $0.345$ & $0.316$ & $0.331$ & $0.231$ & $0.219$ & $0.227$ & $0.230$ \\
    {WocaRDQN+RS} & $0.253$ & $0.222$ & $0.218$ & $0.218$ & $0.218$ & $0.218$ & $0.214$ \\
    {VanillaDQN+RS} & $0.424$ & $0.000$ & $0.000$ & $0.000$ & $0.000$ & $0.000$ & $0.000$ \\
\bottomrule

\end{NiceTabular*}
\end{table}
\begin{table}[t!]
\caption {The average normalized reward of PPO agents under $\ell_{\infty}$ attack. Our S-PPO (\emph{\{base algorithm\}}) constantly achieves a much higher worst reward compared to \emph{\{base algorithm\}}PPO (row (b)) and \emph{\{base algorithm\}}PPO+RS (row (c)), where \emph{\{base algorithm\}} represents various robust training algorithms.}
\label{table:ppo reward}
\centering
\tabcolsep=0.05cm
\tiny
\begin{NiceTabular*}{\linewidth} {@{\extracolsep{\fill}} lccccc|c}[colortbl-like]
\toprule
    \multicolumn{1}{l} {Avg normalized reward} & Clean Reward & MAD & Min-RS & Optimal & PA-AD & Worst Reward\\
    \midrule
    \rowcolor{lightblue} {\textbf{(a) Ours:}} \\
    \rowcolor{lightblue} {S-PPO (SGLD)} & $0.840$ & $\bf{0.837}$ & $\bf{0.745}$ & $0.617$ & $\bf{0.604}$ & $\bf{0.604}$ \\
    \rowcolor{lightblue} {S-PPO (Radial)} & $0.709$ & $0.641$ & $0.263$ & $0.262$ & $0.336$ & $0.262$ \\
    \rowcolor{lightblue} {S-PPO (WocaR)} & $0.745$ & $0.726$ & $0.566$ & $0.531$ & $0.544$ & $0.531$ \\
    \rowcolor{lightblue} {S-PPO (S-ATLA)} & $\bf{0.989}$ & $0.784$ & $0.449$ & $\bf{0.844}$ & $0.553$ & $0.449$ \\
    \rowcolor{lightblue} {S-PPO (S-PA-ATLA)} & $0.935$ & $0.753$ & $0.481$ & $0.234$ & $0.296$ & $0.234$ \\
    \rowcolor{lightblue} {S-PPO (Vanilla)} & $0.929$ & $0.804$ & $0.459$ & $0.226$ & $0.265$ & $0.226$ \\
    \midrule
    \multicolumn{6}{l} {\textbf{(b) SOTA robust agents:}} \\ 
    \multicolumn{1}{l} {SGLDPPO} & $0.800$ & $0.760$ & $0.384$ & $0.418$ & $0.266$ & $0.266$ \\
    \multicolumn{1}{l} {RadialPPO} & $0.658$ & $0.628$ & $0.284$ & $0.133$ & $0.169$ & $0.133$ \\
    \multicolumn{1}{l} {WocaRPPO} & $0.895$ & $0.788$ & $0.342$ & $0.438$ & $0.383$ & $0.342$ \\ 
    \multicolumn{1}{l} {ATLAPPO} & $0.830$ & $0.454$ & $0.232$ & $0.237$ & $0.175$ & $0.175$ \\
    \multicolumn{1}{l} {PA-ATLAPPO} & $0.720$ & $0.609$ & $0.206$ & $0.220$ & $0.274$ & $0.206$ \\
    \multicolumn{1}{l} {VanillaPPO} & $0.870$ & $0.595$ & $0.166$ & $0.136$ & $0.132$ & $0.132$ \\
    \multicolumn{6}{l} {\textbf{(c) Previous smoothed agents:}} \\ 
    \multicolumn{1}{l} {SGLDPPO+RS} & $0.740$ & $0.728$ & $0.420$ & $0.302$ & $0.259$ & $0.259$ \\
    \multicolumn{1}{l} {RadialPPO+RS} & $0.617$ & $0.569$ & $0.195$ & $0.163$ & $0.175$ & $0.163$ \\
    \multicolumn{1}{l} {WocaRPPO+RS} & $0.869$ & $0.797$ & $0.280$ & $0.466$ & $0.336$ & $0.280$ \\
    \multicolumn{1}{l} {ATLAPPO+RS} & $0.847$ & $0.531$ & $0.251$ & $0.263$ & $0.182$ & $0.182$ \\
    \multicolumn{1}{l} {PA-ATLAPPO+RS} & $0.601$ & $0.600$ & $0.224$ & $0.279$ & $0.281$ & $0.224$ \\
    \multicolumn{1}{l} {VanillaPPO+RS} & $0.783$ & $0.585$ & $0.181$ & $0.138$ & $0.151$ & $0.138$\\
\bottomrule
\end{NiceTabular*}
\end{table}

\begin{figure}[!t]
\centering
\vspace{5pt}
\includegraphics[width=0.52\textwidth]{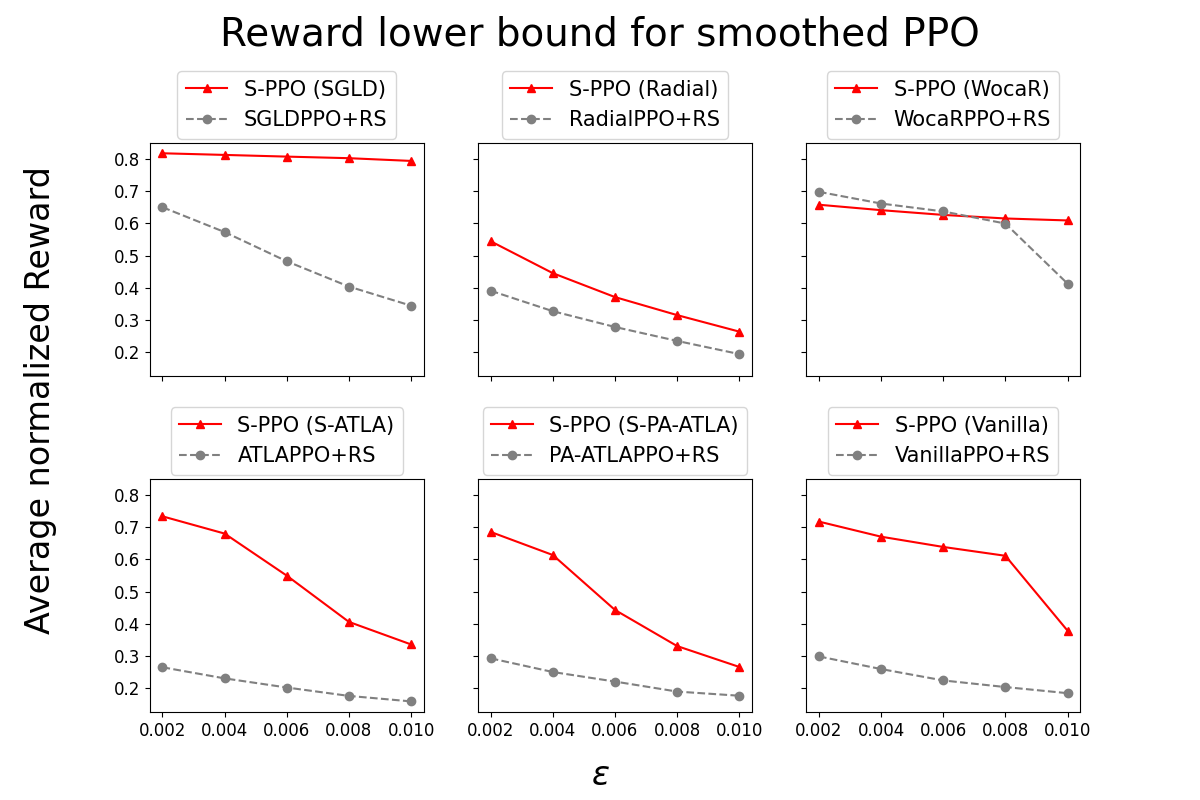}
\vspace{-20pt}
\caption{The certified reward lower bound of smoothed PPO agents. Our S-PPOs demonstrate a much higher lower bound compared to previous smoothed agents.}
\label{fig:lower bound PPO}
\end{figure}
\section{Background and related works}
\label{sec:background}

\begin{table*}[!t]
\vspace{-5pt}
\caption {The comparison between our methods and other DRL agents. Our methods are desirable in both empirical robustness and robustness guarantee.}
\label{table:comparison}
\centering
\tabcolsep=0.05cm
\scriptsize
\begin{NiceTabular*}{\linewidth}{@{\extracolsep{\fill}} lccccc}[colortbl-like]
\toprule
    \multicolumn{1}{l}
    {Methods} & \multicolumn{2}{c} {Empirical Robustness} & \multicolumn{3}{c} {Robustness Guarantee}\\
    \cmidrule(lr){2-3}\cmidrule(lr){4-6}
    {} & Clean Reward$\uparrow$ & Reward under Attack$\uparrow$ & Certified Radius (for DQN) & Action bound (for PPO) & Reward lower bound$\uparrow$ \\
    \midrule
    \rowcolor{lightblue}{\textbf{Our methods:}} \\
    \rowcolor{lightblue}{S-DQN \& S-PPO} & \textbf{Highest} & \textbf{Highest} & \textbf{Yes} & \textbf{Yes} & \textbf{Highest} \\
    \midrule
    {\textbf{SOTA robust agents:}} \\
    {SA-RL \& RADIAL-RL \& WocaR-RL} & High & High & No & No & No\\
    {ATLAPPO \& PA-ATLAPPO (PPO only)} & High & High & No DQN implementation & No & No\\
    {\textbf{Previous smoothed agents:}} \\
    {CROP (DQN only)}& Low & Medium & Yes & No PPO implementation & Low\\
    {Policy Smoothing (PPO only)}& Medium & Medium & No DQN implementation & No derivation & Low\\
\bottomrule
\end{NiceTabular*}
\end{table*}

\paragraph{Randomized smoothing (RS).} Randomized Smoothing \cite{rs} has been proved to provide a robustness guarantee to a \emph{smoothed} classifier under $\ell_2$ perturbation on input examples. The idea is to transform an arbitrary base classifier into an $L$-Lipschitz smoothed classifier by adding Gaussian noises to the input. This transformation facilitates \emph{black-box} robustness verification on the smoothed classifier, which ensures the classification result remains unchanged within the certified radius without the need to know the model parameters. This can be formulated as below. Given a base classifier $f:\mathbb{R}^d\rightarrow\mathcal{Y}$, and let $\Tilde{f}:\mathbb{R}^d\rightarrow\mathcal{Y}$ be the smoothed classifier (i.e., $f$ after RS), $\Tilde{f}$ can be expressed as $\Tilde{f}(x) = \argmax_{c\in\mathcal{Y}}\mathbb{P}_{\delta\sim\mathcal{N}(0,\sigma^2I)}[f(x+\delta)=c]$, where $\delta$ is a random vector following Gaussian distribution $\mathcal{N}(0,\sigma^2I)$. The smoothed classifier $\Tilde{f}$ predicts class $c_A$ with probability $p_A$, and predicts the "runner-up" class $c_B$ with probability $p_B$. The certified radius of $\Tilde{f}$ is denoted as $R$ such that $\Tilde{f}(x+\Delta)=\Tilde{f}(x),\;\forall ||\Delta||_2\leq R$. $R$ can be derived as $R = \dfrac{\sigma}{2}(\Phi^{-1}(p_A)-\Phi^{-1}(p_B))$, where $\Phi^{-1}$ is the inversed Gaussian CDF.
%When we replace $p_A$ and $p_B$ by $\underline{p_A}$ and $\overline{p_B}$, where $\underline{p_A}$ is the lower confidence bound of $p_A$, and $\overline{p_B}$ is the upper confidence bound of $p_B$, the certified radius still holds. In practice, we can use Monte Carlo sampling to estimate $\underline{p_A}$ and $\overline{p_B}$.
%
There have been techniques improving the limitations of RS. For example, \cite{denoiser} proposed to add a denoiser before the original image classifier to remove the Gaussian noises introduced by RS. This approach gives the classifier the ability to tolerate large noises. Our method is the first work leveraging Denoised Smoothing in the DRL setting.

\paragraph{Learning Robust DRL agents.} There are several existing works of learning robust DRL agents through robust training. These agents are non-smoothed DRL agents and their performance is shown in Figure \ref{fig:trade-off} (the grey boxes). \textbf{SA-RL (SADQN and SGLDPPO) \cite{sa}} trained robust agents using a robust regularizer based on the total variation distance and KL-divergence between the perturbed policies and the original policies. \textbf{RADIAL-RL \cite{radialrl}} used the adversarial loss based on the robustness verification bounds as a regularizer. \textbf{WocaR-RL \cite{wocar}} robustify agents through improving the worst-case reward. \textbf{ATLAPPO \cite{optimalattack}} proposed to use the optimal adversary for adversarial training. \textbf{PA-ATLAPPO \cite{paad}} improved ATLA by separating the adversary into a RL-based director and a non-RL actor.

\paragraph{Previous smoothed DRL agents.} Recently, two works proposed to smooth DRL agents in the test-time. \textbf{CROP \cite{crop}} proposed the first framework using RS to study the robustness certification of DRL agents. They showed that the certified radius of a smoothed robustly trained agent is generally larger compared to the smoothed vanilla agents. \textbf{Policy Smoothing \cite{policysmoothing}} demonstrated that the robustness guarantee in the Supervised Learning setting cannot directly transfer to the RL setting due to the non-static nature of RL. They provided an alternative proof for the reward lower bound in the RL setting. However, both approaches perform poorly as shown in Figure \ref{fig:trade-off} (the yellow boxes), suggesting that the previous smoothed agents are not usable in practice, emphasizing the necessity of applying our proposed methods. 

The detailed comparison among our methods, the robust DRL agents, and previous smoothed agents is shown in Table \ref{table:comparison}.
\section{Conclusion and future works}
\label{sec:conclusion}
In this work, we have shown with extensive experiments that our proposed S-DQN and S-PPO agents outperform previous robust agents and smoothed agents in terms of both robustness certificates and robust reward against the current strongest attack, establishing the new state-of-the-art in the field. In future work, we are planning to investigate the idea of leveraging robustness certificates into training to further strengthen the robustness of DRL agents.

\section*{Impact Statement}
This paper investigates certifiably robust Deep Reinforcement Learning (DRL) agents, with a close connection to safety-critical continuous control domains. We hold the belief that our proposed method has the potential to serve as a foundation for constructing more reliable RL tools, thereby positively influencing the broader societal landscape.

\section*{Acknowlegements}
This work is supported in part by National Science Foundation (NSF) awards CNS-1730158, ACI-1540112, ACI-1541349, OAC-1826967, OAC-2112167, CNS-2100237, CNS-2120019, the University of California Office of the President, and the University of California San Diego's California Institute for Telecommunications and Information Technology/Qualcomm Institute. Thanks to CENIC for the 100Gbps networks. C. Sun and T.-W. Weng are supported by National Science Foundation under Grant No. 2107189 and 2313105. T.-W. Weng also thanks the Hellman Fellowship for providing research support. 

\bibliography{ref}
\bibliographystyle{icml2024}

%%%%%%%%%%%%%%%%%%%%%%%%%%%%%%%%%%%%%%%%%%%%%%%%%%%%%%%%%%%%%%%%%%%%%%%%%%%%%%%
%%%%%%%%%%%%%%%%%%%%%%%%%%%%%%%%%%%%%%%%%%%%%%%%%%%%%%%%%%%%%%%%%%%%%%%%%%%%%%%
% APPENDIX
%%%%%%%%%%%%%%%%%%%%%%%%%%%%%%%%%%%%%%%%%%%%%%%%%%%%%%%%%%%%%%%%%%%%%%%%%%%%%%%
%%%%%%%%%%%%%%%%%%%%%%%%%%%%%%%%%%%%%%%%%%%%%%%%%%%%%%%%%%%%%%%%%%%%%%%%%%%%%%%
\newpage
\appendix
\onecolumn

\section{Appendix}
\DoToC
\newpage

\subsection{Detailed algorithms of S-DQN}
\label{sec:Detailed Algorithms of SDQN}

\subsubsection{Training algorithm of S-DQN}
\label{sec:Training Algorithm of SDQN}
The training algorithm of S-DQN is shown in Algorithm \ref{alg:training SDQN}. The algorithm includes all the details of the training procedure introduced in Section \ref{sec:sdqn}. We first add a noise to the current state and take action with $\epsilon$-greedy strategy, Then, store the transitions $\{s_t,a_t,r_t,s_{t+1}\}$ into the replay buffer. Note that the state $s_t$ we stored here is the clean state without noise. When updating the denoiser $D$, we sample a batch of transitions from the replay buffer, add noise to the state again, and compute the loss.
\begin{algorithm}[H]
   \caption{Train S-DQN}
   \label{alg:training SDQN}
\begin{algorithmic}[1]
   \STATE {\bfseries Input:} smoothing variance $\sigma$, steps $T$, replay buffer $\mathcal{B}$, Denoiser $D$, pretrained Q network $Q$
   \FOR{$t=1$ {\bfseries to} $T$}
   \STATE Sample a noise from the normal distribution and add to the state $\tilde{s}_t=s_t+\mathcal{N}(0,\sigma^2I_N)$
   \STATE Select a random action $a_t$ with probability $\epsilon_t$, otherwise $a_t=\argmax_a Q(D(\tilde{s}_t;\theta),a)$
   \STATE Store the transition $\{s_t,a_t,r_t,s_{t+1}\}$ in $\mathcal{B}$
   \STATE Sample a batch of samples $\{s,a,r,s^\prime\}$ from $\mathcal{B}$
   \STATE Sample a noise from the normal distribution and add to the state $\tilde{s}=s+\mathcal{N}(0,\sigma^2I_N)$
   \STATE Compute the reconstruction loss $\mathcal{L}_\textrm{R}=\textrm{MSE}(D(\tilde{s};\theta), s)$
   \STATE Compute the temporal difference loss $\mathcal{L}_{\textrm{TD}}=\textrm{Huber}(r+\gamma\max_{a^\prime}Q(s^\prime,a^\prime)-Q(D(\tilde{s};\theta),a))$
   \STATE Total loss $\mathcal{L}=\lambda_1\mathcal{L}_\textrm{R}+\lambda_2\mathcal{L}_{\textrm{TD}}$
   \STATE Perform gradient descent to minimize loss $\mathcal{L}$ and update the parameters $\theta$ of the denoiser $D$
   \ENDFOR
\end{algorithmic}
\end{algorithm}

\subsubsection{Testing algorithm of S-DQN}
\label{sec:Testing Algorithm of SDQN}
The testing algorithm of S-DQN is shown in Algorithm \ref{alg:test SDQN}. The algorithm includes all the details of the testing procedure introduced in Section \ref{sec:sdqn}. We use the hard randomized smoothing strategy to smooth our agent and do Monte Carlo sampling to estimate the expectation. The definition of $Q_h$ is in Eq.(\ref{e:hard}).
\begin{algorithm}[H]
   \caption{Test S-DQN}
   \label{alg:test SDQN}
\begin{algorithmic}[1]
   \STATE {\bfseries Input:} smoothing variance $\sigma$, number of samples $M$, number of the actions $N$, Denoiser $D$, pretrained Q network $Q$
   \WHILE{not end game}
   \STATE Get state $s$ from the environment
   \FOR{$m=1$ {\bfseries to} $M$}
   \STATE Sample a noise from the normal distribution and add to the state $\tilde{s}_m=s_m+\mathcal{N}(0,\sigma^2I_N)$
   \STATE Store the $Q_h$ value of all the actions $[Q_h(D(\tilde{s}_m), a_1), ..., Q_h(D(\tilde{s}_m), a_N)]$ to the list
   \ENDFOR
   \STATE Take the mean of the $Q_h$ value of each action $\widetilde{Q}(s,a_n)=\frac{1}{M}\Sigma_{m=1}^{M}Q_h(D(\tilde{s}_m), a_n)$
   \STATE Choose the action with the maximum $\widetilde{Q}$ value $a^*=\argmax_{a_n} \widetilde{Q}(s, a_n)$
   \STATE Take action and get the reward
   \ENDWHILE
   \STATE Return the total reward
\end{algorithmic}
\end{algorithm}

\subsubsection{Attack algorithm of smoothed attack}
\label{sec:Attack Algorithm of SDQN}
The algorithm of our Smoothed Attack (S-PGD) is shown in Algorithm \ref{alg:PGD attack designed for SDQN}. The algorithm includes all the details of the attack procedure introduced in Section \ref{sec:sdqn}. Note that our Smoothed Attack considers the noise introduced by randomized smoothing.
\begin{algorithm}[H]
   \caption{Smoothed Attack (S-PGD)}
   \label{alg:PGD attack designed for SDQN}
\begin{algorithmic}[1]
   \STATE {\bfseries Input:} number of iterations $T$, attack budget $\epsilon$, smoothing variance $\sigma$, number of samples $M$, Denoiser $D$, pretrained Q network $Q$
   \STATE Get state $s$ from the environment
   \STATE $\hat{s}=s$
   \FOR{$t=1$ {\bfseries to} $T$}
   \STATE Sample a noise from the normal distribution and add to the state $\tilde{\hat{s}}=\hat{s}+\mathcal{N}(0,\sigma^2I_N)$
   \STATE Compute the cross-entropy loss \\ $\mathcal{L}=-\log\frac{\exp(Q(D(\tilde{\hat{s}}),a^*))}{\Sigma_a\exp(Q(D(\tilde{\hat{s}}),a))}$, \\ where $a^*$ is the original optimal action decided by the agent
   \STATE Calculate the gradient with respect to $\hat{s}$, and project to the $\ell_2$ or $\ell_{\infty}$ norm ball
   \STATE Update $\hat{s}$ by adding the gradient
   \ENDFOR
   \STATE Return the perturbed state $\hat{s}$
\end{algorithmic}
\end{algorithm}
\clearpage

\subsection{Detailed algorithms of S-PPO}
\label{sec:Detailed Algorithms of SPPO}
\subsubsection{Training algorithm of S-PPO}
\label{sec:Training Algorithm of SPPO}
The training algorithm of S-PPO is shown in Algorithm \ref{alg:training SPPO} and \ref{alg:collect trajectories}. The algorithm includes all the details of the training procedure introduced in Section \ref{sec:sppo}. The algorithm of CollectTrajectories function used in step 1 of Algorithm \ref{alg:training SPPO} is shown in Algorithm \ref{alg:collect trajectories}.
\begin{algorithm}[H]
   \caption{Train S-PPO}
   \label{alg:training SPPO}
\begin{algorithmic}[1]
   \STATE {\bfseries Input:} smoothing variance $\sigma$, attack budget $\epsilon$, number of samples $M$, iterations $T$, Policy network $\pi$, Value network $V$
   \FOR{$t=1$ {\bfseries to} $T$}
   \STATE \textbf{// Step 1: Collect trajectories for policy training} \\
   $\{\tau_k\}=$ \textbf{CollectTrajectories()}
   \STATE Compute cumulative reward $\hat{R}_{k,i}$ for each step $i$ in episode $k$ with discount factor $\gamma$
   \STATE \textbf{// Step 2: Update the value network with loss} \\ $\mathcal{L}_V(\theta)=\frac{1}{\Sigma_k|\tau_k|}\Sigma_{\tau_k}\Sigma_i(V(s_{k,i})-\hat{R}_{k,i})^2$
   \STATE \textbf{// Step 3: Update the policy network} 
   \FOR{$m=1$ {\bfseries to} $M$}
   \STATE Sample a noise from the normal distribution and add to the state $\tilde{s}_{k,i,m}=s_{k,i,m}+\mathcal{N}(0,\sigma^2I_N)$
   \STATE Store the output of the policy network $(a^{\textrm{mean}}_{k,i,m},a^{\textrm{std}}_{k,i,m})$ to the list, where $\mathcal{N}(a^{\textrm{mean}}_{k,i,m},a^{\textrm{std}}_{k,i,m})=\pi(a_{k,i,m}|\tilde{s}_{k,i,m})$
   \ENDFOR
   \STATE Take the median and obtain the smoothed policy \\
   $\tilde{\pi}(a_{k,i}|s_{k,i})=\mathcal{N}(\textrm{median}(a^{\textrm{mean}}_{k,i,1}, ..., a^{\textrm{mean}}_{k,i,M}),\textrm{median}(a^{\textrm{std}}_{k,i,1}, ..., a^{\textrm{std}}_{k,i,M}))$
   \STATE Update the policy network with the S-PPO loss \\ 
   $\mathcal{L(\theta)}=-\frac{1}{\Sigma_k|\tau_k|}\Sigma_{\tau_k}\Sigma_i\min(\frac{\tilde{\pi}(a_{k,i}|s_{k,i};\theta)}{\tilde{\pi}(a_{k,i}|s_{k,i};\theta_{\textrm{old}})}\hat{A}_{k,i},\textrm{clip}(\frac{\tilde{\pi}(a_{k,i}|s_{k,i};\theta)}{\tilde{\pi}(a_{k,i}|s_{k,i};\theta_{\textrm{old}})},1-\epsilon_{\textrm{clip}},1+\epsilon_{\textrm{clip}})\hat{A}_{k,i}),$ \\
   where $\hat{A}_{k,i}$ is the advantage
   \ENDFOR
\end{algorithmic}
\end{algorithm}
\begin{algorithm}[H]
   \caption{\textbf{CollectTrajectories} function}
   \label{alg:collect trajectories}
\begin{algorithmic}[1]
   \STATE {\bfseries Input:} number of trajectories $K$, smoothing variance $\sigma$, number of samples $M$, Policy network $\pi$
   \FOR{$k=1$ {\bfseries to} $K$}
   \WHILE{not end game}
   \STATE Get state $s$ from the environment
   \FOR{$m=1$ {\bfseries to} $M$}
   \STATE Sample a noise from the normal distribution and add to the state $\tilde{s}_m=s_m+\mathcal{N}(0,\sigma^2I_N)$
   \STATE Store the mean and standard deviation of the action $(a^{\textrm{mean}}_m,a^{\textrm{std}}_m)$ to the list, where $\mathcal{N}(a^{\textrm{mean}}_m,a^{\textrm{std}}_m)=\pi(a|\tilde{s}_m)$
   \ENDFOR
   \STATE Take the median and obtain the smoothed policy $\tilde{\pi}(a|s)=\mathcal{N}(\textrm{median}(a^{\textrm{mean}}_1, ..., a^{\textrm{mean}}_M),\textrm{median}(a^{\textrm{std}}_1, ..., a^{\textrm{std}}_M))$ \\
   \STATE Take action with the smoothed policy and collect the reward
   \ENDWHILE
   \STATE Store the trajectory $\tau_k$
   \ENDFOR
   \STATE Return the set of the trajectories $\{\tau_{k}\}$
\end{algorithmic}
\end{algorithm}
\clearpage

\subsection{Detailed settings for DQN and PPO}
\label{sec:detailed setting for DQN and PPO}
\subsubsection{Settings for DQN}
\label{sec:settings for DQN}
Our DQN implementation is based on the SADQN \cite{sa} and CROP \cite{crop}. We use the DnCNN structure proposed in \citet{dncnn} as the denoiser to train S-DQN. We train our S-DQN for $300,000$ frames in Pong, Freeway, and RoadRunner. The training time of S-DQN is roughly $12$ hours on our hardware, which is much faster than $40$ hours of SADQN and $17$ hours of RadialDQN. For WocaRDQN, the training is initialized with RadialDQN as we found the training is unstable. The smoothing variance $\sigma$ for S-DQN is set to $0.1$ in Pong, $0.1$ in Freeway, and $0.05$ in RoadRunner. All the experiment results under attack are obtained by taking the average of 5 episodes.
\subsubsection{Settings for PPO}
\label{sec:settings for PPO}
Our PPO implementation is based on the SAPPO \cite{sa}, RadialPPO \cite{radialrl}, ATLAPPO \cite{optimalattack}, and PA-ATLAPPO \cite{paad}. We train S-PPO for $2000000$ steps in Walker and Hopper. We use a simple MLP network for all the PPO algorithms. For the PA-ATLAPPO, we do not combine with SGLD unlike the original paper, as we want to evaluate the true robustness of PA-ATLA algorithm. Note that there is high a variance between the performance of each agent trained with the same algorithm. To get a fair and comparable result, we trained each agent $15$ times and reported the median of the performance as suggested in \citet{sa}. The median agent is selected by considering the median of clean reward, reward under MAD attack, and reward under Min-RS attack from a pool of 15 agents. Subsequently, we conduct further evaluations on the median agents under the Optimal Attack and the PA-AD attack since these evaluations involve high computational costs and are impractical to perform on the entire set of 15 agents. The smoothing variance $\sigma$ for S-PPO is set to $0.2$ in all environments. The $\ell_{\infty}$ attack budget for all the attacks for PPO (MAD, Min-RS, Optimal Attack, PA-AD attack) is set to $0.075$. All the experiment results under attack are obtained by taking the average of 50 episodes.
\clearpage

\subsection{Details of estimating bounds}
\label{sec:detail of estimating bounds}
\subsubsection{Estimating the certified radius for S-DQN}
\label{sec:estimating the CR}
In practice, we use Monte Carlo sampling to estimate $\widetilde{Q}$, which denotes as $\widetilde{Q}_{\textrm{est}}$. The estimation of the Certified Radius is formulated as follows:
\begin{equation}
    R_{\textrm{est},t}=\dfrac{\sigma}{2}(\Phi^{-1}(\widetilde{Q}_{\textrm{est}}(s_t,a_1)-\Delta)-\Phi^{-1}(\widetilde{Q}_{\textrm{est}}(s_t,a_2)+\Delta)),
\end{equation}
where $\widetilde{Q}_{\textrm{est}}(s,a)=\frac{1}{m}\Sigma_{i=1}^mQ_h(D(s+\delta_i),a)$, $\delta_i\sim\mathcal{N}(0,\sigma^2I_N),\forall i\in\{1,...,m\}$, $\Delta=\sqrt{\frac{1}{2m}\ln\frac{1}{\alpha}}$, $m$ is the number of the samples ($m=100$ in our setting), and $\alpha$ is the one-side confidence parameter ($\alpha=0.05$ in our setting). The proof of this estimation can be found in Appendix \ref{sec:CR proof DQN}.

\subsubsection{Estimating the action bound for S-PPO}
\label{sec:estimating the AB}
 In practice, we use Monte Carlo sampling to estimate $\tilde{\pi}_{\textrm{det},p}$, which denotes as $\tilde{\pi}_{\textrm{det},p_{\textrm{est}}}$. The estimation of the Action Bound is formulated as follows:
\begin{equation}
    \begin{split}
        \tilde{\pi}_{\textrm{det},{\underline{p_{\textrm{est}}}}}(s_t)\preceq \tilde{\pi}_{\textrm{det},p_{\textrm{est}}}(s_t+\Delta s)\preceq\tilde{\pi}_{\textrm{det},{\overline{p_{\textrm{est}}}}}(s_t),\;s.t\;||\Delta s||_2\leq\epsilon,
    \end{split}
\end{equation}
where $\tilde{\pi}_{i,\textrm{det},p_{\textrm{est}}}(s)=max\{a_i\in\mathbb{R}|\;\;\lvert\{x\in S_i|x\leq a_i\}\rvert\leq\left\lceil{mp_{\textrm{est}}}\right\rceil\},S_i=\{\pi_{i,\textrm{det}}(s+\delta_1),...,\pi_{i,\textrm{det}}(s+\delta_m)\},\forall i\in\{1,...,N_{\textrm{action}}\}$, $\delta_j\sim\mathcal{N}(0,\sigma^2I_N),\forall j\in\{1,...,m\}$, $\underline{p_{\textrm{est}}}=\Phi(\Phi^{-1}(p_{\textrm{est}}-\Delta)-\frac{\epsilon}{\sigma})$, $\overline{p_{\textrm{est}}}=\Phi(\Phi^{-1}(p_{\textrm{est}}+\Delta)+\frac{\epsilon}{\sigma})$, $\Delta=\sqrt{\frac{1}{2m}\ln\frac{1}{\alpha}}$, $m$ is the number of the samples ($m=100$ in our setting), and $\alpha$ is the one-side confidence parameter ($\alpha=0.05$ in our setting). The proof of this estimation can be found in Appendix \ref{sec:AB proof PPO}.

\subsubsection{Estimating the reward lower bound for smoothed agents}
\label{sec:estimating the reward lower bound}
 In practice, we use Monte Carlo sampling to estimate $\widetilde{F}_{\pi,p}$, which denotes as $\widetilde{F}_{\pi,p_{\textrm{est}}}$. The estimation of the Reward Lower Bound is formulated as follows:
\begin{equation}
    \widetilde{F}_{\pi,p_{\textrm{est}}}(\boldsymbol{\Delta s})\geq\widetilde{F}_{\pi,\underline{p_{\textrm{est}}}}(\boldsymbol{0}),\;\textrm{s.t.}\;||\boldsymbol{\Delta s}||_2\leq B,
\end{equation}
where $\widetilde{F}_{\pi,p_{\textrm{est}}}(\boldsymbol{\Delta s})=max\{r\in\mathbb{R}||\{x\in S|x\leq r\}|\leq\left\lceil{m_\tau p_{\textrm{est}}}\right\rceil\},S=\{F_\pi(\boldsymbol{\delta}_1+\boldsymbol{\Delta s}),...,F_\pi(\boldsymbol{\delta}_{m_\tau}+\boldsymbol{\Delta s})\}$, $\boldsymbol{\delta}_i\sim\mathcal{N}(0,\sigma^2I_{H\times N}),\forall i\in\{1,...,m_\tau\}$, $\underline{p_{\textrm{est}}}=\Phi(\Phi^{-1}(p_{\textrm{est}}-\Delta)-\frac{B}{\sigma})$, $\Delta=\sqrt{\frac{1}{2m_\tau}\ln\frac{1}{\alpha}}$, $m_\tau$ is the number of sample trajectories ($m_\tau=1000$ in our setting), and $\alpha$ is the one-side confidence parameter ($\alpha=0.05$ in our setting). Note that in this setting, each state is added with a noise. Therefore, $m=1$. The proof of this estimation can be found in Appendix \ref{sec:reward lower bound proof}.
\clearpage

\subsection{Proof of the certified radius for S-DQN}
\label{sec:CR proof DQN}
In this section, we give the formal proof of the certified radius introduced in Section \ref{sec:certification}. Our proof is based on the proof proposed by \citet{smoothadv} in Appendix A. Recall that we have:
\begin{equation}
\label{e:CR appendix}
    R_t=\dfrac{\sigma}{2}(\Phi^{-1}(\widetilde{Q}(s_t,a_1))-\Phi^{-1}(\widetilde{Q}(s_t,a_2))),
\end{equation}
where $a_1$ is the action with the largest Q-value among all the other actions, $a_2$ is the ”runner-up” action, $R_t$ is the certified radius at time $t$, $\Phi$ is the CDF of normal distribution, $\sigma$ is the smoothing variance, and $\widetilde{Q}(s,a)$ is defined in Eq.(\ref{e:11}).

We first go over the lemma needed for proof.
\paragraph{Lemma 1} For the function $Q_h:\mathcal{S}\times\mathcal{A}\rightarrow [0,1]$, the function $\widetilde{Q}$ is $\dfrac{1}{\sigma}\sqrt{\dfrac{2}{\pi}}$-Lipschitz.

\emph{Proof.} From the definition of $\widetilde{Q}$, we have
\begin{equation}
    \widetilde{Q}(s,a)=(Q_h\ast\mathcal{N}(0,\sigma^2 I_n))(D(s),a)=\dfrac{1}{(2\pi)^{\sfrac{n}{2}}\sigma^n}\int_{\mathbb{R}_n}Q_h(D(t),a)\exp\Bigl(-\dfrac{1}{2\sigma^2}||s-t||_2^2\Bigl)dt.
\end{equation}
Take the gradient w.r.t. s, we have
\begin{equation}
    \nabla_s\widetilde{Q}(s,a)=\dfrac{1}{(2\pi)^{\sfrac{n}{2}}\sigma^n}\int_{\mathbb{R}_n}\dfrac{1}{\sigma^2}(s-t)Q_h(D(t),a)\exp\Bigl(-\dfrac{1}{2\sigma^2}||s-t||_2^2\Bigl)dt.
\end{equation}
For any unit direction $u$, we have
\begin{equation}
    \begin{split}
        u\cdot\nabla_s\widetilde{Q}(s,a) & \leq\dfrac{1}{(2\pi)^{\sfrac{n}{2}}\sigma^n}\int_{\mathbb{R}_n}\dfrac{1}{\sigma^2}|u\cdot(s-t)|\exp\Bigl(-\dfrac{1}{2\sigma^2}||s-t||_2^2\Bigl)dt \\
        & =\dfrac{1}{\sigma^2}\int_{\mathbb{R}_n}\dfrac{1}{\sqrt{2\pi}\sigma}|u\cdot(s-t)|\exp\Bigl(-\dfrac{1}{2\sigma^2}||s-t||_2^2\Bigl)dt \\
        & =\dfrac{1}{\sigma^2}\int_{-\infty}^{+\infty}\dfrac{1}{\sqrt{2\pi}\sigma}|z|\exp\Bigl(-\dfrac{1}{2\sigma^2}z^2\Bigl)dz \\
        & =\dfrac{1}{\sigma^2}\mathbb{E}_{z\sim\mathcal{N}(0,\sigma^2)}[|z|] \\
        & =\dfrac{1}{\sigma}\sqrt{\dfrac{2}{\pi}}.
    \end{split}
\end{equation}
In fact, there is a stronger smoothness property for $\widetilde{Q}$.
\paragraph{Lemma 2} For the function $Q_h:\mathcal{S}\times\mathcal{A}\rightarrow [0,1]$, the mapping $s\mapsto\sigma\Phi^{-1}(\widetilde{Q}(s,a))$ is $1$-Lipschitz.

\emph{Proof.} Take the gradient of $\Phi^{-1}(\widetilde{Q}(s,a))$ w.r.t. s, we have
\begin{equation}
    \nabla\Phi^{-1}(\widetilde{Q}(s,a))=\dfrac{\nabla\widetilde{Q}(s,a)}{\Phi^\prime(\Phi^{-1}(\widetilde{Q}(s,a)))}.
\end{equation}
We intend to show that for any unit direction $u$,
\begin{equation}
    \begin{gathered}
        u\cdot\sigma\nabla\Phi^{-1}(\widetilde{Q}(s,a))\leq 1 \\
        u\cdot\sigma\nabla\widetilde{Q}(s,a)\leq \Phi^\prime(\Phi^{-1}(\widetilde{Q}(s,a))) \\
        u\cdot\sigma\nabla\widetilde{Q}(s,a)\leq \dfrac{1}{\sqrt{2\pi}}\exp\Bigl(-\dfrac{1}{2}(\Phi^{-1}(\widetilde{Q}(s,a)))^2\Bigl).
    \end{gathered}
\end{equation}
The left-hand side can be written as
\begin{equation}
    \dfrac{1}{\sigma}\mathbb{E}_{\delta\sim\mathcal{N}(0,\sigma^2I_n)}[Q_h(D(s+\delta),a)\delta\cdot u].
\end{equation}
We claim that the supremum of the above quantity over all functions $Q_h:\mathcal{S}\times\mathcal{A}\rightarrow [0,1]$, subject to $\mathbb{E}[Q_h(D(s+\delta),a)]=\widetilde{Q}(s,a)$, is equal to
\begin{equation}
    \dfrac{1}{\sigma}\mathbb{E}[(\delta\cdot u)\mathds{1}\{\delta\cdot u\geq -\sigma\Phi^{-1}(\widetilde{Q}(s,a))\}]=\dfrac{1}{\sqrt{2\pi}}\exp\Bigl(-\dfrac{1}{2}(\Phi^{-1}(\widetilde{Q}(s,a)))^2\Bigl).
\end{equation}

To prove the claim is true, note that $h:\delta\mapsto\mathds{1}\{\delta\cdot u\geq -\sigma\Phi^{-1}(\widetilde{Q}(s,a))\}$ achieves equality. Assume by contradiction that the maximum is reached by some function $f:\delta\rightarrow [0,1]$. Consider the set $\Omega^+=\{\delta|h(\delta)>f(\delta)\}$ and the set $\Omega^-=\{\delta|h(\delta)<f(\delta)\}$. Now construct the new function $f^\prime=f+(h-f)\mathds{1}\{\Omega^+\}-(f-h)\mathds{1}\{\Omega^-\}$, which takes value in $[0,1]$. Since both $h$ and $f$ integrate to $\widetilde{Q}(s,a)$, we have $\int_{\Omega^+}(h-f)d\delta=\int_{\Omega^-}(f-h)d\delta$. This gives that $f^\prime$ also integrates to $\widetilde{Q}(s,a)$. By the definition of $h$, for any $\delta_1\in\Omega^+$ and $\delta_2\in\Omega^-$, we have $\delta_1\cdot u>\delta_2\cdot u$, and since $\int_{\Omega^+}(h-f)d\delta=\int_{\Omega^-}(f-h)d\delta$, we have
\begin{equation}
    \begin{gathered}
        \int_{\Omega^+}(\delta\cdot u)(h-f)(\delta)d\delta>\int_{\Omega^-}(\delta\cdot u)(f-h)(\delta)d\delta \\
        \int(\delta\cdot u)f(\delta)d\delta<\int(\delta\cdot u)f(\delta)d\delta+\int_{\Omega^+}(\delta\cdot u)(h-f)(\delta)d\delta-\int_{\Omega^-}(\delta\cdot u)(f-h)(\delta)d\delta \\
        \int(\delta\cdot u)f(\delta)d\delta<\int(\delta\cdot u)f^\prime(\delta)d\delta \\
    \end{gathered}
\end{equation}
Hence, the maximum is obtained at $h$. The claim holds, and hence, we have
\begin{equation}
    u\cdot\sigma\nabla\Phi^{-1}(\widetilde{Q}(s,a))\leq 1.
\end{equation}
Now, we can prove the certified radius in Eq.(\ref{e:CR appendix}).
\paragraph{Theorem 1} Let $Q_h:\mathcal{S}\times\mathcal{A}\rightarrow[0,1]$, and $\widetilde{Q}(s,a)=\mathbb{E}_{\delta\sim\mathcal{N}(0,\sigma^2I)}Q_h(D(s+\delta),a)$. At time step $t$ with state $s_t$, the certified radius is
\begin{equation}
    R_t=\dfrac{\sigma}{2}(\Phi^{-1}(\widetilde{Q}(s_t,a_1))-\Phi^{-1}(\widetilde{Q}(s_t,a_2))),
\end{equation}
where $a_1$ is the action with the largest Q-value among all the other actions, $a_2$ is the ”runner-up” action, $R_t$ is the certified radius at time $t$, $\Phi$ is the CDF of normal distribution, and $\sigma$ is the smoothing variance. The certified radius gives a lower bound on the minimum $\ell_2$ adversarial perturbation required to change the policy from $a_1$ to $a_2$.

\emph{Proof.} Let the perturbation be $\Delta s$ and able to change the action from $a_1$ to $a_2$. By lemma 2, we have
\begin{equation}
    \sigma\Phi^{-1}(\widetilde{Q}(s_t,a_1))-\sigma\Phi^{-1}(\widetilde{Q}(s_t+\Delta s,a_1))\leq ||\Delta s||_2
\end{equation}
Since the perturbation can change the action, we have $\widetilde{Q}(s_t+\Delta s,a_1)\leq\widetilde{Q}(s_t+\Delta s,a_2)$, which leads to
\begin{equation}
\label{e:a}
    \sigma\Phi^{-1}(\widetilde{Q}(s_t,a_1))-\sigma\Phi^{-1}(\widetilde{Q}(s_t+\Delta s,a_2))\leq ||\Delta s||_2
\end{equation}
By lemma 2 and $\widetilde{Q}(s_t+\Delta s,a_2)\geq\widetilde{Q}(s_t,a_2)$, we have
\begin{equation}
\label{e:b}
    \sigma\Phi^{-1}(\widetilde{Q}(s_t+\Delta s,a_2))-\sigma\Phi^{-1}(\widetilde{Q}(s_t,a_2))\leq ||\Delta s||_2
\end{equation}
Combine Eq.(\ref{e:a}) and Eq.(\ref{e:b}), we have
\begin{equation}
    ||\Delta s||_2\geq\dfrac{\sigma}{2}(\Phi^{-1}(\widetilde{Q}(s_t,a_1))-\Phi^{-1}(\widetilde{Q}(s_t,a_2))),
\end{equation}
which gives us the certified radius
\begin{equation}
    R_t=\dfrac{\sigma}{2}(\Phi^{-1}(\widetilde{Q}(s_t,a_1))-\Phi^{-1}(\widetilde{Q}(s_t,a_2))).
\end{equation}

Now, we prove the practical version of the certified radius introduced in Appendix \ref{sec:estimating the CR}:
\paragraph{Theorem 2} Let $Q_h:\mathcal{S}\times\mathcal{A}\rightarrow[0,1]$, and $\widetilde{Q}_{\textrm{est}}(s,a)=\frac{1}{m}\Sigma_{i=1}^mQ_h(D(s+\delta_i),a),\delta_i\sim\mathcal{N}(0,\sigma^2I_N)$, $\forall i\in\{1,...,m\}$. At time step $t$ with state $s_t$, the certified radius is
\begin{equation}
    R_{\textrm{est},t}=\dfrac{\sigma}{2}(\Phi^{-1}(\widetilde{Q}_{\textrm{est}}(s_t,a_1)-\Delta)-\Phi^{-1}(\widetilde{Q}_{\textrm{est}}(s_t,a_2)+\Delta)),
\end{equation}
where $\Delta=\sqrt{\frac{1}{2m}\ln\frac{1}{\alpha}}$, $m$ is the number of the samples, $\alpha$ is the one-side confidence parameter, $a_1$ is the action with the largest Q-value among all the other actions, $a_2$ is the ”runner-up” action, $R_t$ is the certified radius at time $t$, $\Phi$ is the CDF of normal distribution, and $\sigma$ is the smoothing variance.

\emph{Proof.} By \emph{Hoeffding’s Inequality}, for any $t\geq 0$, we have
\begin{equation}
    P(\widetilde{Q}_{\textrm{est}}-\widetilde{Q}\geq t) \leq \exp^{-2mt^2}. 
\end{equation}
Rearrange the inequality
\begin{equation}
    P(\widetilde{Q}_{\textrm{est}}-\widetilde{Q}\geq \sqrt{\frac{1}{2m}\ln\frac{1}{\alpha}}) \leq \alpha.   
\end{equation}
Hence, a $1-\alpha$ confidence lower bound $\underline{\widetilde{Q}}$ of $\widetilde{Q}$ is
\begin{equation}
    \underline{\widetilde{Q}}=\widetilde{Q}_{\textrm{est}}-\sqrt{\frac{1}{2m}\ln\frac{1}{\alpha}}=\widetilde{Q}_{\textrm{est}}-\Delta.
\end{equation}
Similarly, we have $1-\alpha$ confidence upper bound $\overline{\widetilde{Q}}$ of $\widetilde{Q}$
\begin{equation}
    \overline{\widetilde{Q}}=\widetilde{Q}_{\textrm{est}}+\Delta.
\end{equation}
Substitute $\widetilde{Q}(s_t,a_1)$ with the lower bound and $\widetilde{Q}(s_t,a_2)$ with the upper bound, we have
\begin{equation}
    R_{\textrm{est},t}=\dfrac{\sigma}{2}(\Phi^{-1}(\widetilde{Q}_{\textrm{est}}(s_t,a_1)-\Delta)-\Phi^{-1}(\widetilde{Q}_{\textrm{est}}(s_t,a_2)+\Delta))
\end{equation}

\clearpage
\subsection{Proof of the action bound for S-PPO}
\label{sec:AB proof PPO}
In this section, we give the formal proof of the action bound introduced in Section \ref{sec:certification}. Our proof is based on the proof proposed by \citet{median} in Appendix B. Recall that we have:
\begin{equation}
    \begin{split}
        & \tilde{\pi}_{\textrm{det},{\underline{p}}}(s_t) \preceq \tilde{\pi}_{\textrm{det},p}(s_t+\Delta s)\preceq\tilde{\pi}_{\textrm{det},{\overline{p}}}(s_t),\;s.t\;||\Delta s||_2\leq\epsilon,
    \end{split}
\end{equation}
where $\tilde{\pi}_{i,\textrm{det},p}(s)=\textrm{sup}\{a_i\in\mathbb{R}|\mathbb{P}_{\delta\sim\mathcal{N}(0,\sigma^2I)}[\pi_{i,\textrm{det}}(s+\delta)\leq a_i]\leq p\},\forall i\in\{1,...,N_{\textrm{action}}\}$, $\underline{p}=\Phi(\Phi^{-1}(p)-\frac{\epsilon}{\sigma})$, $\overline{p}=\Phi(\Phi^{-1}(p)+\frac{\epsilon}{\sigma})$, $\Phi$ is the CDF of normal distribution, and $\sigma$ is the smoothing variance.
\paragraph{Theorem 3} Let $\pi:\mathcal{S}\rightarrow\mathcal{A}$ be the policy network, and $\tilde{\pi}_{i,\textrm{det},p}(s)=\textrm{sup}\{a_i\in\mathbb{R}|\mathbb{P}_{\delta\sim\mathcal{N}(0,\sigma^2I)}[\pi_{i,\textrm{det}}(s+\delta)\leq a_i]\leq p\},\forall i\in\{1,...,N_{\textrm{action}}\}$. At time step $t$ with state $s_t$, the action bound is
\begin{equation}
    \begin{split}
        & \tilde{\pi}_{\textrm{det},{\underline{p}}}(s_t) \preceq \tilde{\pi}_{\textrm{det},p}(s_t+\Delta s)\preceq\tilde{\pi}_{\textrm{det},{\overline{p}}}(s_t),\;s.t\;||\Delta s||_2\leq\epsilon,
    \end{split}
\end{equation}
where $\underline{p}=\Phi(\Phi^{-1}(p)-\frac{\epsilon}{\sigma})$, $\overline{p}=\Phi(\Phi^{-1}(p)+\frac{\epsilon}{\sigma})$, $\Phi$ is the CDF of a normal distribution, and $\sigma$ is the smoothing variance.

\emph{Proof.} Let $\mathcal{E}_i(s_t)=\mathbb{E}_{\delta\sim\mathcal{N}(0,\sigma^2I_N)}[\mathds{1}\{\pi_{i,\textrm{det}}(s_t+\delta)\leq\tilde{\pi}_{i,\textrm{det},{\underline{p}}}(s_t)\}]$, and we have $\mathcal{E}_i:\mathbb{R}^N\rightarrow[0,1]$, $\forall i\in\{1,...,N_{\textrm{action}}\}$. The mapping $s_t\mapsto\sigma\Phi^{-1}(\mathcal{E}_i(s_t))$ is $1$-Lipschitz, which can be proved by the similar technique used in Lemma 2. Since $\mathcal{E}_i(s_t)=\mathbb{P}_{\delta\sim\mathcal{N}(0,\sigma^2I_N)}[\pi_{i,\textrm{det}}(s_t+\delta)\leq\tilde{\pi}_{i,\textrm{det},{\underline{p}}}(s_t)]$, given the perturbation $\Delta s$, we have
\begin{equation}
    \begin{split}
        & \sigma\Phi^{-1}(\mathbb{P}_{\delta\sim\mathcal{N}(0,\sigma^2I_N)}[\pi_{i,\textrm{det}}(s_t+\delta+\Delta s)\leq\tilde{\pi}_{i,\textrm{det},{\underline{p}}}(s_t)])- \\
        & \sigma\Phi^{-1}(\mathbb{P}_{\delta\sim\mathcal{N}(0,\sigma^2I_N)}[\pi_{i,\textrm{det}}(s_t+\delta)\leq\tilde{\pi}_{i,\textrm{det},{\underline{p}}}(s_t)])\leq||\Delta s||_2.
    \end{split}
\end{equation}
Rearrange the inequality, we have
\begin{equation}
    \begin{split}
        & \Phi^{-1}(\mathbb{P}_{\delta\sim\mathcal{N}(0,\sigma^2I_N)}[\pi_{i,\textrm{det}}(s_t+\delta+\Delta s)\leq\tilde{\pi}_{i,\textrm{det},{\underline{p}}}(s_t)]) \\
        & \leq\Phi^{-1}(\mathbb{P}_{\delta\sim\mathcal{N}(0,\sigma^2I_N)}[\pi_{i,\textrm{det}}(s_t+\delta)\leq\tilde{\pi}_{i,\textrm{det},{\underline{p}}}(s_t)])+\dfrac{||\Delta s||_2}{\sigma} \\
        & \leq\Phi^{-1}(\mathbb{P}_{\delta\sim\mathcal{N}(0,\sigma^2I_N)}[\pi_{i,\textrm{det}}(s_t+\delta)\leq\tilde{\pi}_{i,\textrm{det},{\underline{p}}}(s_t)])+\dfrac{\epsilon}{\sigma} \\
        & =\Phi^{-1}(\underline{p})+\dfrac{\epsilon}{\sigma} \\
        & =\Phi^{-1}(p).
    \end{split}
\end{equation}
By the monotonicity of $\Phi$, we have
\begin{equation}
    \mathbb{P}_{\delta\sim\mathcal{N}(0,\sigma^2I_N)}[\pi_{i,\textrm{det}}(s_t+\delta+\Delta s)\leq\tilde{\pi}_{i,\textrm{det},{\underline{p}}}(s_t)]\leq p.
\end{equation}
Recall that $\tilde{\pi}_{i,\textrm{det},p}(s_t+\Delta s)=\textrm{sup}\{a_i\in\mathbb{R}|\mathbb{P}_{\delta\sim\mathcal{N}(0,\sigma^2I_N)}[\pi_{i,\textrm{det}}(s_t+\delta+\Delta s)\leq a_i]\leq p\},\forall i\in\{1,...,N_{\textrm{action}}\}$, we have
\begin{equation}
    \tilde{\pi}_{\textrm{det},{\underline{p}}}(s_t)\preceq\tilde{\pi}_{\textrm{det},p}(s_t+\Delta s).
\end{equation}
We can show that $    \tilde{\pi}_{\textrm{det},p}(s_t+\Delta s)\preceq\tilde{\pi}_{\textrm{det},{\overline{p}}}(s_t)$ for all $||\Delta s||_2\leq\epsilon$ with the similar technique. Combine the two bounds we have
\begin{equation}
    \tilde{\pi}_{\textrm{det},{\underline{p}}}(s_t)\preceq\tilde{\pi}_{\textrm{det},p}(s_t+\Delta s)\preceq\tilde{\pi}_{\textrm{det},{\overline{p}}}(s_t).
\end{equation}
Now, we prove the practical version of the action bound introduced in Appendix \ref{sec:estimating the AB}:
\paragraph{Theorem 4} Let $\pi:\mathcal{S}\rightarrow\mathcal{A}$ be the policy network, and $\tilde{\pi}_{i,\textrm{det},p_{\textrm{est}}}(s)=max\{a_i\in\mathbb{R}|\;\;\lvert\{x\in S_i|x\leq a_i\}\rvert\leq\left\lceil{mp_{\textrm{est}}}\right\rceil\},S_i=\{\pi_{i,\textrm{det}}(s+\delta_1),...,\pi_{i,\textrm{det}}(s+\delta_m)\},\forall i\in\{1,...,N_{\textrm{action}}\}$, $\delta_j\sim\mathcal{N}(0,\sigma^2I_N),\forall j=1,...,m$. At time step $t$ with state $s_t$, the action bound is
\begin{equation}
    \begin{split}
        \tilde{\pi}_{\textrm{det},{\underline{p_{\textrm{est}}}}}(s_t)\preceq \tilde{\pi}_{\textrm{det},p_{\textrm{est}}}(s_t+\Delta s)\preceq\tilde{\pi}_{\textrm{det},{\overline{p_{\textrm{est}}}}}(s_t),\;s.t\;||\Delta s||_2\leq\epsilon,
    \end{split}
\end{equation}
where $\underline{p_{\textrm{est}}}=\Phi(\Phi^{-1}(p_{\textrm{est}}-\Delta)-\frac{\epsilon}{\sigma})$, $\overline{p_{\textrm{est}}}=\Phi(\Phi^{-1}(p_{\textrm{est}}+\Delta)+\frac{\epsilon}{\sigma})$, $\Delta=\sqrt{\frac{1}{2m}\ln\frac{1}{\alpha}}$, $m$ is the number of the samples, $\alpha$ is the one-side confidence parameter, $\Phi$ is the CDF of normal distribution, and $\sigma$ is the smoothing variance.

\emph{Proof.} By \emph{Hoeffding’s Inequality}, for any $t\geq 0$, we have
\begin{equation}
    P(p_{\textrm{est}}-p\geq t) \leq \exp^{-2mt^2}. 
\end{equation}
Rearrange the inequality
\begin{equation}
    P(p_{\textrm{est}}-p\geq \sqrt{\frac{1}{2m}\ln\frac{1}{\alpha}}) \leq \alpha.   
\end{equation}
Hence, a $1-\alpha$ confidence lower bound $\underline{p}$ of $p$ is
\begin{equation}
    \underline{p}=p_{\textrm{est}}-\sqrt{\frac{1}{2m}\ln\frac{1}{\alpha}}=p_{\textrm{est}}-\Delta.
\end{equation}
Similarly, we have $1-\alpha$ confidence upper bound $\overline{p}$ of $\underline{p}$
\begin{equation}
    \overline{p}=p_{\textrm{est}}+\Delta.
\end{equation}
Substitute $\Phi(\Phi^{-1}(p)-\frac{\epsilon}{\sigma})$ with the lower bound, and $\Phi(\Phi^{-1}(p)+\frac{\epsilon}{\sigma})$ with the upper bound, we have
\begin{equation}
    \big[\Phi(\Phi^{-1}(p_{\textrm{est}}-\Delta)-\frac{\epsilon}{\sigma}),\;\;\Phi(\Phi^{-1}(p_{\textrm{est}}+\Delta)+\frac{\epsilon}{\sigma}\big],
\end{equation}
which is the new upper bound and lower bound in the expression.
\clearpage
\subsection{Proof of the reward lower bound for smoothed agents}
\label{sec:reward lower bound proof}
In this section, we give the formal proof of the reward lower bound introduced in Section \ref{sec:certification}. Our proof is based on the proof proposed by \citet{median} in Appendix B. Recall that we have:
\begin{equation}
    \widetilde{F}_{\pi,p}(\boldsymbol{\Delta s})\geq\widetilde{F}_{\pi,\underline{p}}(\boldsymbol{0}),\;\textrm{s.t.}\;||\boldsymbol{\Delta s}||_2\leq B,
\end{equation}
where $\widetilde{F}_{\pi,p}(\boldsymbol{\Delta s})=\textrm{sup}\{r\in\mathbb{R}|\mathbb{P}_{\boldsymbol{\delta}\sim\mathcal{N}(0,\sigma^2I_{H\times N})}[F_\pi(\boldsymbol{\delta}+\boldsymbol{\Delta s})\leq r]\leq p\}$, $\underline{p}=\Phi(\Phi^{-1}(p)-\frac{B}{\sigma})$, and $B$ is the $\ell_2$ attack budget of the entire trajectory.
\paragraph{Theorem 5} Let $F_\pi :\mathbb{R}^{H\times N}\rightarrow\mathbb{R}$ be the function mapping the perturbation to the total reward, and $\widetilde{F}_{\pi,p}(\boldsymbol{\Delta s})=\textrm{sup}\{r\in\mathbb{R}|\mathbb{P}_{\boldsymbol{\delta}\sim\mathcal{N}(0,\sigma^2I_{H\times N})}[F_\pi(\boldsymbol{\delta}+\boldsymbol{\Delta s})\leq r]\leq p\}$. The reward lower bound is
\begin{equation}
    \widetilde{F}_{\pi,p}(\boldsymbol{\Delta s})\geq\widetilde{F}_{\pi,\underline{p}}(\boldsymbol{0}),\;\textrm{s.t.}\;||\boldsymbol{\Delta s}||_2\leq B,
\end{equation}
where $\underline{p}=\Phi(\Phi^{-1}(p)-\frac{B}{\sigma})$, $B$ is the $\ell_2$ attack budget of the entire trajectory, $\Phi$ is the CDF of normal distribution, and $\sigma$ is the smoothing variance.

\emph{Proof.} Let $\mathcal{E}(\boldsymbol{\Delta s})=\mathbb{E}_{\delta\sim\mathcal{N}(0,\sigma^2I_{H\times N})}[\mathds{1}\{F_\pi(\boldsymbol{\delta}+\boldsymbol{\Delta s})\leq\widetilde{F}_{\pi,\underline{p}}(\boldsymbol{0})\}]$, and we have $\mathcal{E}:\mathbb{R}^{H\times N}\rightarrow[0,1]$. The mapping $\boldsymbol{\Delta s}\mapsto\sigma\Phi^{-1}(\mathcal{E}(\boldsymbol{\Delta s}))$ is $1$-Lipschitz by Lemma 2. Since $\mathcal{E}(\boldsymbol{\Delta s})=\mathbb{P}_{\delta\sim\mathcal{N}(0,\sigma^2I_{H\times N})}[F_\pi(\boldsymbol{\delta}+\boldsymbol{\Delta s})\leq\widetilde{F}_{\pi,\underline{p}}(\boldsymbol{0})]$, given the perturbation $\boldsymbol{\Delta s}$, we have
\begin{equation}
    \begin{split}
        & \sigma\Phi^{-1}(\mathbb{P}_{\delta\sim\mathcal{N}(0,\sigma^2I_{H\times N})}[F_\pi(\boldsymbol{\delta}+\boldsymbol{\Delta s})\leq\widetilde{F}_{\pi,\underline{p}}(\boldsymbol{0})])-\sigma\Phi^{-1}(\mathbb{P}_{\delta\sim\mathcal{N}(0,\sigma^2I_{H\times N})}[F_\pi(\boldsymbol{\delta})\leq\widetilde{F}_{\pi,\underline{p}}(\boldsymbol{0})]) \\
        & \leq||\boldsymbol{\Delta s}||_2.
    \end{split}
\end{equation}
Rearrange the inequality, we have
\begin{equation}
    \begin{split}
        & \Phi^{-1}(\mathbb{P}_{\delta\sim\mathcal{N}(0,\sigma^2I_{H\times N})}[F_\pi(\boldsymbol{\delta}+\boldsymbol{\Delta s})\leq\widetilde{F}_{\pi,\underline{p}}(\boldsymbol{0})]) \\
        & \leq\Phi^{-1}(\mathbb{P}_{\delta\sim\mathcal{N}(0,\sigma^2I_{H\times N})}[F_\pi(\boldsymbol{\delta})\leq\widetilde{F}_{\pi,\underline{p}}(\boldsymbol{0})])+\dfrac{||\boldsymbol{\Delta s}||_2}{\sigma} \\
        & \leq\Phi^{-1}(\mathbb{P}_{\delta\sim\mathcal{N}(0,\sigma^2I_{H\times N})}[F_\pi(\boldsymbol{\delta})\leq\widetilde{F}_{\pi,\underline{p}}(\boldsymbol{0})])+\dfrac{B}{\sigma} \\
        & =\Phi^{-1}(\underline{p})+\dfrac{B}{\sigma} \\
        & =\Phi^{-1}(p).
    \end{split}
\end{equation}
By the monotonicity of $\Phi$, we have
\begin{equation}
    \mathbb{P}_{\delta\sim\mathcal{N}(0,\sigma^2I_{H\times N})}[F_\pi(\boldsymbol{\delta}+\boldsymbol{\Delta s})\leq\widetilde{F}_{\pi,\underline{p}}(\boldsymbol{0})]\leq p.
\end{equation}
Recall that $\widetilde{F}_{\pi,p}(\boldsymbol{\Delta s})=\textrm{sup}\{r\in\mathbb{R}|\mathbb{P}_{\boldsymbol{\delta}\sim\mathcal{N}(0,\sigma^2I_{H\times N})}[F_\pi(\boldsymbol{\delta}+\boldsymbol{\Delta s})\leq r]\leq p\}$, we have
\begin{equation}
    \widetilde{F}_{\pi,p}(\boldsymbol{\Delta s})\geq\widetilde{F}_{\pi,\underline{p}}(\boldsymbol{0}).
\end{equation}
Now, we prove the practical version of the reward lower bound introduced in Appendix \ref{sec:estimating the reward lower bound}:
\paragraph{Theorem 6} Let $F_\pi :\mathbb{R}^{H\times N}\rightarrow\mathbb{R}$ be the function mapping the perturbation to the total reward, and $\widetilde{F}_{\pi,p_{\textrm{est}}}(\boldsymbol{\Delta s})=max\{r\in\mathbb{R}||\{x\in S|x\leq r\}|\leq\left\lceil{m_\tau p_{\textrm{est}}}\right\rceil\},S=\{F_\pi(\boldsymbol{\delta}_1+\boldsymbol{\Delta s}),...,F_\pi(\boldsymbol{\delta}_{m_\tau}+\boldsymbol{\Delta s})\}$, $\boldsymbol{\delta}_i\sim\mathcal{N}(0,\sigma^2I_{H\times N}),\forall i=\{1,...,m_\tau\}$. The reward lower bound is
\begin{equation}
    \widetilde{F}_{\pi,p_{\textrm{est}}}(\boldsymbol{\Delta s})\geq\widetilde{F}_{\pi,\underline{p_{\textrm{est}}}}(\boldsymbol{0}),\;\textrm{s.t.}\;||\boldsymbol{\Delta s}||_2\leq B,
\end{equation}
where $\underline{p_{\textrm{est}}}=\Phi(\Phi^{-1}(p_{\textrm{est}}-\Delta)-\frac{B}{\sigma})$, $\Delta=\sqrt{\frac{1}{2m_\tau}\ln\frac{1}{\alpha}}$, $m_\tau$ is the number of sample trajectories, $\alpha$ is the one-side confidence parameter, $\Phi$ is the CDF of normal distribution, and $\sigma$ is the smoothing variance.

\emph{Proof.} By \emph{Hoeffding’s Inequality}, for any $t\geq 0$, we have
\begin{equation}
    P(p_{\textrm{est}}-p\geq t) \leq \exp^{-2m_\tau t^2}. 
\end{equation}
Rearrange the inequality
\begin{equation}
    P(p_{\textrm{est}}-p\geq \sqrt{\frac{1}{2m_\tau}\ln\frac{1}{\alpha}}) \leq \alpha.   
\end{equation}
Hence, a $1-\alpha$ confidence lower bound $\underline{p}$ of $p$ is
\begin{equation}
    \underline{p}=p_{\textrm{est}}-\sqrt{\frac{1}{2m_\tau}\ln\frac{1}{\alpha}}=p_{\textrm{est}}-\Delta.
\end{equation}
Substitute $\Phi(\Phi^{-1}(p)-\frac{B}{\sigma})$ with the lower bound, we have
\begin{equation}
    \Phi(\Phi^{-1}(p_{\textrm{est}}-\Delta)-\frac{B}{\sigma}),
\end{equation}
which is the new lower bound in the expression.
\clearpage

\subsection{The certified radius of smoothed DQN agents}
\label{sec:cr result}
Table \ref{table:cr} presents the Certified Radius of our S-DQNs and CROP's agents. Our S-DQN agents generally achieve higher Certified Radius. It's important to note that while the CROP framework used a sample number of $m=10000$ for estimating the Certified Radius, we used $m=100$ here. Although a larger $m$ can enhance confidence in estimating and result in a larger Certified Radius, $m=10000$ is not a practical setting. Our hard randomized smoothing strategy demonstrates the capability to provide a large Certified Radius even with a small $m$.
\begin{table}[H]
\caption {The Certified Radius of different smoothed DQN agents.}
\label{table:cr}
\centering
\tabcolsep=0.05cm
\begin{tabular*}{\linewidth} {@{\extracolsep{\fill}} lccc}
\toprule
    \multicolumn{1}{l}
    {Methods} & \multicolumn{3}{c} {Certified Radius (larger is better)} \\
    \cmidrule(lr){2-4}
    {} & Pong & Freeway & RoadRunner \\
    \midrule
    {\textbf{Ours (using hard randomized smoothing):}} \\
    \multicolumn{1}{l} {S-DQN (Radial)} & $\bf{0.1044}$ & $\bf{0.1134}$ & $\bf{0.0576}$ \\
    \multicolumn{1}{l} {S-DQN (S-PGD)} & $0.0502$ & $0.0766$ & $0.0520$ \\
    \multicolumn{1}{l} {S-DQN (Vanilla)} & $0.0619$ & $0.0774$ & $0.0502$ \\
    {\textbf{CROP \cite{crop} (using mean smoothing):}} \\
    \multicolumn{1}{l} {RadialDQN+RS} & $0.0000$ & $0.0000$ & $0.0000$ \\
    \multicolumn{1}{l} {SADQN+RS} & $0.0615$ & $0.0665$ & $0.0000$ \\
    \multicolumn{1}{l} {VanillaDQN+RS} & $0.0000$ & $0.0000$ & $0.0000$ \\
\bottomrule
\end{tabular*}
\end{table}
\clearpage

\subsection{The Action Divergence of smoothed PPO agents}
\label{sec:adiv result}
We designed a metric based on the action bound in Section \ref{sec:certification} to evaluate the certified robustness of the smoothed PPO agents. We define the \textbf{Action Divergence} as follows:
\begin{equation}
   \textrm{ADIV}=\mathbb{E}_{s,\epsilon}\Bigl[\dfrac{||\tilde{\pi}_{\textrm{det},{\overline{p_{\textrm{est}}}}}(s)-\tilde{\pi}_{\textrm{det},{\underline{p_{\textrm{est}}}}}(s)||_2}{2\epsilon}\Bigl],
\end{equation}
where $\epsilon$ is the $\ell_2$ attack budget used in estimating the action bound, and the definition of $\overline{p_{\textrm{est}}}$ and $\underline{p_{\textrm{est}}}$ is in Appendix \ref{sec:estimating the AB}. We found that the $\ell_2$ norm of the difference between the upper and lower bound of the actions is proportional to the magnitude of the $\ell_2$ budget $\epsilon$, which makes $\frac{||\tilde{\pi}_{\textrm{det},{\overline{p_{\textrm{est}}}}}(s)-\tilde{\pi}_{\textrm{det},{\underline{p_{\textrm{est}}}}}(s)||_2}{2\epsilon}$ almost unchanged under different $\epsilon$ setting. Hence, we take the expectation over the state $s$ and the budget $\epsilon$ to estimate this fraction, which is the ADIV proposed here. We estimate the ADIV by taking the average of 50 trajectories with three different $\epsilon$ settings ($\epsilon=0.1$, $\epsilon=0.2$, and $\epsilon=0.3$).

ADIV describes the worst-case stability of the actions of a smoothed PPO agent under any $\ell_2$ perturbation. The more this value is, the more unstable the smoothed agent is under the $\ell_2$ attack. The result is shown in Table \ref{table:adiv}. Our S-PPO agents exhibit lower ADIV compared to their naively smoothed counterparts. Notably, S-PPO (SGLD) and S-PPO (WocaR) have the lowest ADIV, and they also demonstrate higher robustness under attacks compared to the others in our study.
\begin{table}[H]
\caption {The Action Divergence of different smoothed PPO agents.}
\label{table:adiv}
\centering
\tabcolsep=0.05cm
\begin{tabular*}{\linewidth} {@{\extracolsep{\fill}} lcc}
\toprule
    \multicolumn{1}{l}
    {Methods} & \multicolumn{2}{c} {Action Divergence (lower is better)} \\
    \cmidrule(lr){2-3}
    {} & Walker & Hopper\\
    \midrule
    {\textbf{Ours:}} \\
    \multicolumn{1}{l} {S-PPO (SGLD)} & $1.401$ & $\bf{0.656}$ \\
    \multicolumn{1}{l} {S-PPO (Radial)} & $8.665$ & $2.305$ \\
    \multicolumn{1}{l} {S-PPO (WocaR)} & $\bf{1.125}$ & $0.778$ \\
    \multicolumn{1}{l} {S-PPO (S-ATLA)} & $4.218$ & $8.964$ \\
    \multicolumn{1}{l} {S-PPO (S-PA-ATLA)} & $3.899$ & $8.432$ \\
    \multicolumn{1}{l} {S-PPO (Vanilla)} & $2.926$ & $1.618$ \\
    {\textbf{Previous smoothed agents:}} \\
    \multicolumn{1}{l} {SGLDPPO+RS} & $2.221$ & $1.375$ \\
    \multicolumn{1}{l} {RadialPPO+RS} & $7.964$ & $2.766$ \\
    \multicolumn{1}{l} {WocaRPPO+RS} & $2.431$ & $1.466$ \\
    \multicolumn{1}{l} {ATLAPPO+RS} & $6.062$ & $16.994$ \\
    \multicolumn{1}{l} {PA-ATLAPPO+RS} & $5.595$ & $11.165$ \\
    \multicolumn{1}{l} {VanillaPPO+RS} & $5.030$ & $4.187$ \\
\bottomrule
\end{tabular*}
\end{table}
\clearpage

\subsection{Detailed experiment results of robust reward for S-DQN}
\label{sec: detailed robust reward for S-DQN}
Table \ref{table:detailed dqn reward} shows the reward of DQN agents under $\ell_{\infty}$ PGD attack and PA-AD attack. Note that we used our stronger S-PGD attack and S-PA-AD to evaluate all the smoothed agents. Our S-DQN (Vanilla) already outperformed the state-of-the-art robust agent, RadialDQN, in most of the settings except for RoadRunner. This problem was solved by introducing S-DQN (Radial) and S-DQN (S-PGD). S-DQN (Radial) performs especially well under all attacks across various environments, which suggests that our S-DQN can be further boosted by changing the base model to a robust agent.
\begin{table}[H]
\caption {The reward of DQN agents under  $\ell_{\infty}$ PGD attack and PA-AD attack. The smoothing variance $\sigma$ for the smoothed agents is set to $0.1$ in Pong, $0.1$ in Freeway, and $\sigma=0.05$ in RoadRunner.}
\label{table:detailed dqn reward}
\tabcolsep=0.015cm
\centering
\footnotesize
\begin{tabular*}{1\linewidth} {@{\extracolsep{\fill}} lccccccc}
\toprule
    {Pong} 
    & \multicolumn{1}{c} {Clean reward} & \multicolumn{5}{c} {PGD or S-PGD} & \multicolumn{1}{c} {PAAD or S-PAAD}\\
    \cmidrule(lr){3-7}
    {$\epsilon(\ell_{\infty})$} &  & $0.01$ & $0.02$ & $0.03$ & $0.04$ & $0.05$ & $0.05$\\
    \midrule
    {\textbf{Ours:}} \\
    {S-DQN (Radial)} & $\bf{21.0\!\pm\!0.0}$ & $\bf{21.0\!\pm\!0.0}$ & $\bf{21.0\!\pm\!0.0}$ & $\bf{21.0\!\pm\!0.0}$ &
    $\bf{21.0\!\pm\!0.0}$ & $\bf{20.8\!\pm\!0.4}$ & $14.0\!\pm\!2.1$ \\
    {S-DQN (S-PGD)} & $20.6\!\pm\!0.5$ & $20.8\!\pm\!0.4$ & $20.0\!\pm\!1.1$ & $15.6\!\pm\!4.3$ & $13.8\!\pm\!4.8$ & $1.6\!\pm\!4.2$ & $11.0\!\pm\!2.6$ \\
    {S-DQN (Vanilla)} & $20.4\!\pm\!0.5$ & $\bf{21.0\!\pm\!0.0}$ & $20.4\!\pm\!0.8$ & $20.2\!\pm\!0.8$ & $16.6\!\pm\!4.4$ & $18.4\!\pm\!2.1$ & $\bf{18.6\!\pm\!1.2}$ \\
    \multicolumn{2}{l} {\textbf{SOTA robust agents:}} \\
    {RadialDQN} & $\bf{21.0\!\pm\!0.0}$ & $\bf{21.0\!\pm\!0.0}$ & $20.0\!\pm\!2.0$ & $-20.2\!\pm\!0.4$ & $-20.6\!\pm\!0.5$ & $-21.0\!\pm\!0.00$ & $-21.0\!\pm\!0.00$ \\
    {SADQN} & $\bf{21.0\!\pm\!0.0}$ & $\bf{21.0\!\pm\!0.0}$ & $-19.4\!\pm\!0.8$ & $-21.0\!\pm\!0.0$ & $-21.0\!\pm\!0.0$ & $-21.0\!\pm\!0.0$ & $-21.0\!\pm\!0.0$ \\
    {WocaRDQN} & $20.0\!\pm\!0.9$ & $19.6\!\pm\!1.4$ & $-20.4\!\pm\!0.8$ & $-20.8\!\pm\!0.4$ & $-21.0\!\pm\!0.00$ & $-21.0\!\pm\!0.00$ & $-21.0\!\pm\!0.00$ \\
    {VanillaDQN} & $\bf{21.0\!\pm\!0.0}$ & $-21.0\!\pm\!0.0$ & $-21.0\!\pm\!0.0$ & $-21.0\!\pm\!0.0$ & $-21.0\!\pm\!0.0$ & $-21.0\!\pm\!0.0$ & $-21.0\!\pm\!0.0$ \\
    \multicolumn{4}{l} {\textbf{Previous smoothed agents:}} \\
    {RadialDQN+RS} & $-21.0\!\pm\!0.0$ & $-21.0\!\pm\!0.0$ & $-21.0\!\pm\!0.0$ & $-21.0\!\pm\!0.0$ & $-21.0\!\pm\!0.0$ & $-21.0\!\pm\!0.0$ & $-21.0\!\pm\!0.0$ \\
    {SADQN+RS} & $-21.0\!\pm\!0.0$ & $-21.0\!\pm\!0.0$ & $-21.0\!\pm\!0.0$ & $-21.0\!\pm\!0.0$ & $-21.0\!\pm\!0.0$ & $-21.0\!\pm\!0.0$ & $-21.0\!\pm\!0.0$ \\
    {WocaRDQN+RS} & $-21.0\!\pm\!0.0$ & $-21.0\!\pm\!0.0$ & $-21.0\!\pm\!0.0$ & $-21.0\!\pm\!0.0$ & $-21.0\!\pm\!0.0$ & $-21.0\!\pm\!0.0$ & $-21.0\!\pm\!0.0$ \\
    {VanillaDQN+RS} & $-20.8\!\pm\!0.4$ & $-21.0\!\pm\!0.0$ & $-21.0\!\pm\!0.0$ & $-21.0\!\pm\!0.0$ & $-21.0\!\pm\!0.0$ & $-21.0\!\pm\!0.0$ & $-21.0\!\pm\!0.0$ \\
\toprule
    {Freeway} \\
    \midrule
    {\textbf{Ours:}} \\
    {S-DQN (Radial)} & $33.0\!\pm\!0.0$ & $\bf{33.0\!\pm\!0.0}$ & $\bf{32.6\!\pm\!0.5}$ & $\bf{32.6\!\pm\!0.5}$ & $\bf{31.6\!\pm\!0.5}$ & $\bf{32.0\!\pm\!1.1}$ & $\bf{32.0\!\pm\!1.1}$ \\
    {S-DQN (S-PGD)} & $32.6\!\pm\!1.4$ & $32.6\!\pm\!1.0$ & $32.0\!\pm\!1.3$ & $30.2\!\pm\!0.8$ & $28.2\!\pm\!1.5$ & $25.6\!\pm\!0.5$ & $30.4\!\pm\!1.0$ \\
    {S-DQN (Vanilla)} & $\bf{34.0\!\pm\!0.0}$ & $\bf{33.0\!\pm\!0.9}$ & $31.4\!\pm\!1.0$ & $28.0\!\pm\!1.4$ & $20.4\!\pm\!1.9$ & $6.6\!\pm\!2.2$ & $13.0\!\pm\!2.1$ \\
    \multicolumn{2}{l} {\textbf{SOTA robust agents:}} \\
    {RadialDQN} & $32.6\!\pm\!0.5$ & $\bf{33.0\!\pm\!0.0}$ & $28.4\!\pm\!1.2$ & $22.8\!\pm\!1.9$ & $20.0\!\pm\!1.1$ & $21.0\!\pm\!0.6$ & $22.8\!\pm\!1.7$ \\
    {SADQN} & $30.0\!\pm\!0.0$ & $30.0\!\pm\!0.0$ & $27.2\!\pm\!1.2$ & $20.4\!\pm\!0.5$ & $20.8\!\pm\!1.0$ & $18.8\!\pm\!1.3$ & $21.0\!\pm\!1.8$ \\
    {WocaRDQN} & $32.2\!\pm\!1.2$ & $29.0\!\pm\!1.3$ & $21.8\!\pm\!2.0$ & $20.6\!\pm\!0.8$ & $21.2\!\pm\!1.0$ & $22.0\!\pm\!0.0$ & $21.4\!\pm\!1.6$ \\
    {VanillaDQN} & $\bf{34.0\!\pm\!0.0}$ & $0.0\!\pm\!0.0$ & $0.0\!\pm\!0.0$ & $0.0\!\pm\!0.0$ & $0.0\!\pm\!0.0$ & $0.0\!\pm\!0.0$ & $0.0\!\pm\!0.0$ \\
    \multicolumn{4}{l} {\textbf{Previous smoothed agents:}} \\
    {RadialDQN+RS} & $22.2\!\pm\!2.2$ & $22.2\!\pm\!2.2$ & $22.2\!\pm\!2.2$ & $22.2\!\pm\!2.2$ & $22.2\!\pm\!2.2$ & $22.2\!\pm\!2.2$ & $21.8\!\pm\!1.2$ \\
    {SADQN+RS} & $22.2\!\pm\!2.2$ & $22.2\!\pm\!2.2$ & $22.2\!\pm\!2.2$ & $22.2\!\pm\!2.9$ & $21.6\!\pm\!1.7$ & $22.8\!\pm\!0.8$ & $22.6\!\pm\!1.2$ \\
    {WocaRDQN+RS} & $22.2\!\pm\!2.2$ & $22.2\!\pm\!2.2$ & $22.2\!\pm\!2.2$ & $22.2\!\pm\!2.2$ & $22.2\!\pm\!2.2$ & $22.2\!\pm\!2.2$ & $21.8\!\pm\!1.2$ \\
    {VanillaDQN+RS} & $22.2\!\pm\!2.2$ & $0.0\!\pm\!0.0$ & $0.0\!\pm\!0.0$ & $0.0\!\pm\!0.0$ & $0.0\!\pm\!0.0$ & $0.0\!\pm\!0.0$ & $0.0\!\pm\!0.0$ \\
\toprule
    {RoadRunner} \\
    \midrule
    {\textbf{Ours:}} \\
    {S-DQN (Radial)} & $39380\!\pm\!4579$ & $39360\!\pm\!4566$ & $\bf{40480\!\pm\!8076}$ & $25640\!\pm\!3232$ & $21060\!\pm\!2286$ & $\bf{13020\!\pm\!4935}$ & $\bf{11220\!\pm\!4324}$ \\
    {S-DQN (S-PGD)} & $42780\!\pm\!6316$ & $42620\!\pm\!3953$ & $35740\!\pm\!5420$ & $\bf{27380\!\pm\!8896}$ & $\bf{21360\!\pm\!9340}$ & $2840\!\pm\!1756$ & $0\!\pm\!0$ \\
    {S-DQN (Vanilla)} & $47480\!\pm\!8807$ & $23320\!\pm\!3932$ & $3460\!\pm\!5924$ & $0\!\pm\!0$ & $0\!\pm\!0$ & $0\!\pm\!0$ & $0\!\pm\!0$ \\
    \multicolumn{2}{l} {\textbf{SOTA robust agents:}} \\
    {RadialDQN} & $39620\!\pm\!4821$ & $\bf{43520\!\pm\!4081}$ & $24160\!\pm\!2604$ & $15500\!\pm\!6466$ & $1020\!\pm\!937$ & $620\!\pm\!492$ & $3560\!\pm\!488$ \\
    {SADQN} & $46680\!\pm\!7742$ & $28580\!\pm\!2584$ & $3240\!\pm\!1544$ & $780\!\pm\!840$ & $420\!\pm\!523$ & $100\!\pm\!200$ & $2640\!\pm\!1317$ \\
    {WocaRDQN} & $32480\!\pm\!5096$ & $1580\!\pm\!2108$ & $0\!\pm\!0$ & $0\!\pm\!0$ & $0\!\pm\!0$ & $0\!\pm\!0$ & $20\!\pm\!40$ \\
    {VanillaDQN} & $\bf{48320\!\pm\!5989}$& $0\!\pm\!0$ & $0\!\pm\!0$ & $0\!\pm\!0$ & $0\!\pm\!0$ & $0\!\pm\!0$ & $0\!\pm\!0$ \\
    \multicolumn{4}{l} {\textbf{Previous smoothed agents:}} \\
    {RadialDQN+RS} & $13420\!\pm\!1955$ & $11260\!\pm\!2504$ & $9220\!\pm\!3080$ & $6680\!\pm\!1705$ & $7780\!\pm\!1900$ & $3180\!\pm\!1326$ & $7420\!\pm\!2604$ \\
    {SADQN+RS} & $18520\!\pm\!2510$ & $14240\!\pm\!6013$ & $16440\!\pm\!3817$ & $1960\!\pm\!1323$ & $1040\!\pm\!1074$ & $560\!\pm\!973$ & $1180\!\pm\!922$ \\
    {WocaRDQN+RS} & $5120\!\pm\!3319$ & $560\!\pm\!647$ & $0\!\pm\!0$ & $0\!\pm\!0$ & $0\!\pm\!0$ & $0\!\pm\!0$ & $0\!\pm\!0$ \\
    {VanillaDQN+RS} & $29640\!\pm\!5271$ & $0\!\pm\!0$ & $0\!\pm\!0$ & $0\!\pm\!0$ & $0\!\pm\!0$ & $0\!\pm\!0$ & $0\!\pm\!0$ \\
\bottomrule

\end{tabular*}
\end{table}
\clearpage

\subsection{Detailed experiment results of reward lower bound for S-DQN}
\label{sec: detailed reward lower bound for DQN}
Table \ref{table:reward lower bound DQN} shows the details of the reward lower bound for smoothed DQN agents under different $\ell_2$ budget $\epsilon$. We use the same budget $\epsilon$ for every state, and hence, the total budget is $B=\epsilon\sqrt{H}$, where $H$ is the length of the trajectory. We set $H=2500$ in Pong, Freeway, and RoadRunner. The reward lower bound of S-DQN (Vanilla) is comparable with the bound of S-DQN (Radial) and S-DQN (S-PGD), suggesting that our S-DQN already achieves a high robustness guarantee without further combining with other robust agents or leveraging adversarial training.
\begin{table}[H]
\caption {The reward lower bound of smoothed DQN agents under different $\ell_2$ attack budgets. The smoothing variance $\sigma$ for all the agents is set to $0.1$ in Pong, $0.1$ in Freeway, and $\sigma=0.05$ in RoadRunner.}
\label{table:reward lower bound DQN}
\centering
\begin{tabular*}
{1\linewidth} {@{\extracolsep{\fill}} lccccc}
\toprule
    {Pong} 
    & \multicolumn{5}{c} {$\ell_2$ attack budget} \\
    \cmidrule(lr){2-6}
    {$\epsilon(\ell_2)$} & $0.001$ & $0.002$ & $0.003$ & $0.004$ & $0.005$ \\
    \midrule
    {\textbf{Ours:}} \\
    {S-DQN (Radial)} & $\bf{20.0}$ & $\bf{20.0}$ & $\bf{19.0}$ & $\bf{18.0}$ & $\bf{18.0}$ \\
    {S-DQN (S-PGD)} & $18.0$ & $17.0$ & $16.0$ & $14.0$ & $11.0$ \\
    {S-DQN (Vanilla)} & $18.0$ & $17.0$ & $16.0$ & $15.0$ & $14.0$ \\
    {\textbf{Previous smoothed agents:}} \\
    {RadialDQN+RS} & $-21.0$ & $-21.0$ & $-21.0$ & $-21.0$ & $-21.0$ \\
    {SADQN+RS} & $-21.0$ & $-21.0$ & $-21.0$ & $-21.0$ & $-21.0$ \\
    {WocaRDQN+RS} & $-21.0$ & $-21.0$ & $-21.0$ & $-21.0$ & $-21.0$ \\
    {VanillaDQN+RS} & $-21.0$ & $-21.0$ & $-21.0$ & $-21.0$ & $-21.0$ \\
\toprule
    {Freeway} & \multicolumn{5}{c} {}\\
    \midrule
    {\textbf{Ours:}} \\
    {S-DQN (Radial)} & $\bf{31.0}$ & $\bf{30.0}$ & $\bf{29.0}$ & $28.0$ & $\bf{28.0}$ \\
    {S-DQN (S-PGD)} & $\bf{31.0}$ & $\bf{30.0}$ & $\bf{29.0}$ & $28.0$ & $27.0$ \\
    {S-DQN (Vanilla)} & $\bf{31.0}$ & $\bf{30.0}$ & $\bf{29.0}$ & $\bf{29.0}$ & $\bf{28.0}$ \\
    {\textbf{Previous smoothed agents:}} \\
    {RadialDQN+RS} & $20.0$ & $20.0$ & $20.0$ & $20.0$ & $19.0$ \\
    {SADQN+RS} & $20.0$ & $20.0$ & $20.0$ & $20.0$ & $19.0$ \\
    {WocaRDQN+RS} & $20.0$ & $20.0$ & $20.0$ & $20.0$ & $19.0$ \\
    {VanillaDQN+RS} & $13.0$ & $12.0$ & $11.0$ & $10.0$ & $9.0$ \\
\toprule
    {RoadRunner} & \multicolumn{5}{c} {}\\
    \midrule
    {\textbf{Ours:}} \\
    {S-DQN (Radial)} & $\bf{36200}$ & $\bf{29400}$ & $\bf{21612}$ & $14163$ & $14001$ \\
    {S-DQN (S-PGD)} & $33000$ & $24483$ & $19295$ & $\bf{19104}$ & $\bf{19100}$ \\
    {S-DQN (Vanilla)} & $32215$ & $25097$ & $21123$ & $18067$ & $18001$ \\
    {\textbf{Previous smoothed agents:}} \\
    {RadialDQN+RS} & $9400$ & $5497$ & $2295$ & $2104$ & $2100$ \\
    {SADQN+RS} & $17900$ & $15197$ & $12623$ & $9567$ & $9501$ \\
    {WocaRDQN+RS} & $3300$ & $1200$ & $593$ & $306$ & $300$ \\
    {VanillaDQN+RS} & $500$ & $100$ & $0$ & $0$ & $0$ \\
\bottomrule

\end{tabular*}
\end{table}
\clearpage

\subsection{Detailed experiment results of robust reward for S-PPO}
\label{sec: detailed robust reward for S-PPO}
Table \ref{table:detailed ppo reward} shows the reward of PPO agents under different $\ell_{\infty}$ attacks. Note that we trained each agent $15$ times and reported the median of the performance as suggested in \citet{sa} to get a fair and comparable result. Our S-PPO exhibits high clean reward and robust reward under attacks in all environments, while the previous smoothed agents only achieve similar performance compared to the original robust agents.
\begin{table}[H]
\caption {The reward of PPO agents under different attacks. The smoothing variance $\sigma$ for all the smoothed agents is set to $0.2$ in Walker and Hopper. The $\ell_{\infty}$ attack budget is set to $0.075$ in both environments.}
\label{table:detailed ppo reward}
\centering
\tabcolsep=0.05cm
\footnotesize
\begin{tabular*}{\linewidth} {@{\extracolsep{\fill}} lccccc}
\toprule
    \multicolumn{1}{l}
    {Walker} & Clean reward & MAD attack & Min-RS attack & Optimal attack & PA-AD attack \\
    \midrule
    \multicolumn{1}{l} {\textbf{Ours:}} \\
    \multicolumn{1}{l} {S-PPO (SGLD)} & $4566$ & $\bf{4537}$ & $\bf{4241}$ & $4085$ & $\bf{4026}$ \\
    \multicolumn{1}{l} {S-PPO (Radial)} & $2117$ & $2160$ & $1028$ & $915$ & $689$ \\
    \multicolumn{1}{l} {S-PPO (WocaR)} & $4363$ & $4360$ & $3907$ & $3920$ & $3867$ \\
    \multicolumn{1}{l} {S-PPO (S-ATLA)} & $\bf{4897}$ & $4460$ & $2170$ & $\bf{5010}$ & $2980$ \\
    \multicolumn{1}{l} {S-PPO (S-PA-ATLA)} & $4407$ & $4045$ & $2379$ & $144$ & $372$ \\
    \multicolumn{1}{l} {S-PPO (Vanilla)} & $4552$ & $4386$ & $3203$ & $944$ & $1077$ \\
    \multicolumn{2}{l} {\textbf{SOTA robust agents:}} \\
    \multicolumn{1}{l} {SGLDPPO} & $4329$ & $4177$ & $2376$ & $2747$ & $718$ \\
    \multicolumn{1}{l} {RadialPPO} & $2221$ & $2230$ & $1270$ & $132$ & $152$ \\
    \multicolumn{1}{l} {WocaRPPO} & $4110$ & $3918$ & $1950$ & $2916$ & $2067$ \\
    \multicolumn{1}{l} {ATLAPPO} & $3564$ & $2567$ & $672$ & $818$ & $263$ \\
    \multicolumn{1}{l} {PA-ATLAPPO} & $2548$ & $1717$ & $591$ & $183$ & $298$ \\
    \multicolumn{1}{l} {VanillaPPO} & $4301$ & $2806$ & $551$ & $437$ & $275$ \\
    \multicolumn{4}{l} {\textbf{Previous smoothed agents:}} \\
    \multicolumn{1}{l} {SGLDPPO+RS} & $4290$ & $4124$ & $2739$ & $1615$ & $717$ \\
    \multicolumn{1}{l} {RadialPPO+RS} & $1804$ & $1883$ & $610$ & $145$ & $208$ \\
    \multicolumn{1}{l} {WocaRPPO+RS} & $4013$ & $4160$ & $1362$ & $3211$ & $1765$ \\
    \multicolumn{1}{l} {ATLAPPO+RS} & $4129$ & $3348$ & $894$ & $1090$ & $322$ \\
    \multicolumn{1}{l} {PA-ATLAPPO+RS} & $1325$ & $1990$ & $427$ & $322$ & $332$ \\
    \multicolumn{1}{l} {VanillaPPO+RS} & $3582$ & $2892$ & $592$ & $440$ & $401$ \\
\toprule
    \multicolumn{1}{l}
    {Hopper} \\
    \midrule
    \multicolumn{1}{l} {\textbf{Ours:}} \\
    \multicolumn{1}{l} {S-PPO (SGLD)} & $2894$ & $2896$ & $\bf{2428}$ & $1579$ & $1523$ \\
    \multicolumn{1}{l} {S-PPO (Radial)} & $3756$ & $\bf{3205}$ & $1212$ & $1285$ & $\bf{2015}$ \\
    \multicolumn{1}{l} {S-PPO (WocaR)} & $2335$ & $2194$ & $1328$ & $1053$ & $1189$ \\
    \multicolumn{1}{l} {S-PPO (S-ATLA)} & $\bf{3770}$ & $2557$ & $1752$ & $\bf{2595}$ & $1927$ \\
    \multicolumn{1}{l} {S-PPO (S-PA-ATLA)} & $3737$ & $2631$ & $1839$ & $1655$ & $1950$ \\
    \multicolumn{1}{l} {S-PPO (Vanilla)} & $3583$ & $2765$ & $1049$ & $995$ & $1190$ \\
    \multicolumn{2}{l} {\textbf{SOTA robust agents:}} \\
    \multicolumn{1}{l} {SGLDPPO} & $2772$ & $2587$ & $1107$ & $1087$ & $1463$ \\
    \multicolumn{1}{l} {RadialPPO} & $3291$ & $3056$ & $1182$ & $900$ & $1161$ \\
    \multicolumn{1}{l} {WocaRPPO} & $3652$ & $2993$ & $1111$ & $1112$ & $1331$ \\
    \multicolumn{1}{l} {ATLAPPO} & $3577$ & $1493$ & $1245$ & $1172$ & $1124$ \\
    \multicolumn{1}{l} {PA-ATLAPPO} & $3508$ & $3297$ & $1110$ & $1518$ & $1842$ \\
    \multicolumn{1}{l} {VanillaPPO} & $3321$ & $2375$ & $834$ & $695$ & $789$ \\
    \multicolumn{4}{l} {\textbf{Previous smoothed agents:}} \\
    \multicolumn{1}{l} {SGLDPPO+RS} & $2354$ & $2386$ & $1106$ & $1059$ & $1411$ \\
    \multicolumn{1}{l} {RadialPPO+RS} & $3298$ & $2876$ & $1011$ & $1122$ & $1165$ \\
    \multicolumn{1}{l} {WocaRPPO+RS} & $3535$ & $2878$ & $1084$ & $1095$ & $1208$ \\
    \multicolumn{1}{l} {ATLAPPO+RS} & $3278$ & $1485$ & $1220$ & $1161$ & $1129$ \\
    \multicolumn{1}{l} {PA-ATLAPPO+RS} & $3537$ & $3027$ & $1365$ & $1861$ & $1866$ \\
    \multicolumn{1}{l} {VanillaPPO+RS} & $3211$ & $2238$ & $920$ & $707$ & $840$ \\
\bottomrule
\end{tabular*}
\end{table}
\clearpage

\subsection{Detailed experiment results of reward lower bound for S-PPO}
\label{sec: detailed reward lower bound for PPO}
Table \ref{table:reward lower bound PPO} shows the details of the reward lower bound for smoothed PPO agents under different $\ell_2$ budget $\epsilon$. We use the same budget $\epsilon$ for every state, and hence, the total budget $B=\epsilon\sqrt{H}$, where $H$ is the length of the trajectory. We set $H=1000$ in Walker and Hopper. Our S-PPOs exhibit higher reward lower bounds compared to their naively smoothed counterparts.
\begin{table}[H]
\caption {The reward lower bound of smoothed PPO agents under different $\ell_2$ attack budgets. The smoothing variance $\sigma$ for all the agents is set to $0.2$ in all environments.}
\label{table:reward lower bound PPO}
\centering
\begin{tabular*}
{1\linewidth} {@{\extracolsep{\fill}} lccccc}
\toprule
    {Walker} 
    & \multicolumn{5}{c} {$\ell_2$ attack budget} \\
    \cmidrule(lr){2-6}
    {$\epsilon(\ell_2)$} & $0.002$ & $0.004$ & $0.006$ & $0.008$ & $0.01$ \\
    \midrule
    \multicolumn{1}{l} {\textbf{Ours:}} \\
    \multicolumn{1}{l} {S-PPO (SGLD)} & $4496$ & $4478$ & $4460$ & $\bf{4440}$ & $\bf{4420}$ \\
    \multicolumn{1}{l} {S-PPO (Radial)} & $1648$ & $1413$ & $1159$ & $817$ & $550$ \\
    \multicolumn{1}{l} {S-PPO (WocaR)} & $4345$ & $4333$ & $4322$ & $4308$ & $4296$ \\
    \multicolumn{1}{l} {S-PPO (S-ATLA)} & $\bf{4781}$ & $\bf{4556}$ & $3571$ & $2287$ & $1746$ \\
    \multicolumn{1}{l} {S-PPO (S-PA-ATLA)} & $4364$ & $4017$ & $2526$ & $1573$ & $1100$ \\
    \multicolumn{1}{l} {S-PPO (Vanilla)} & $4585$ & $4531$ & $\bf{4476}$ & $4368$ & $2189$ \\
    \multicolumn{4}{l} {\textbf{Previous smoothed agents:}} \\
    \multicolumn{1}{l} {SGLDPPO+RS} & $4159$ & $3703$ & $2886$ & $2236$ & $1839$ \\
    \multicolumn{1}{l} {RadialPPO+RS} & $1160$ & $987$ & $821$ & $654$ & $420$ \\
    \multicolumn{1}{l} {WocaRPPO+RS} & $4235$ & $4195$ & $4130$ & $3969$ & $2178$ \\
    \multicolumn{1}{l} {ATLAPPO+RS} & $935$ & $735$ & $568$ & $378$ & $307$ \\
    \multicolumn{1}{l} {PA-ATLAPPO+RS} & $606$ & $512$ & $455$ & $416$ & $385$ \\
    \multicolumn{1}{l} {VanillaPPO+RS} & $1263$ & $979$ & $853$ & $748$ & $657$ \\
\toprule
    {Hopper} \\
    \midrule
    \multicolumn{1}{l} {\textbf{Ours:}} \\
    \multicolumn{1}{l} {S-PPO (SGLD)} & $2783$ & $\bf{2758}$ & $\bf{2732}$ & $\bf{2710}$ & $\bf{2661}$ \\
    \multicolumn{1}{l} {S-PPO (Radial)} & $\bf{2865}$ & $2294$ & $1925$ & $1760$ & $1574$ \\
    \multicolumn{1}{l} {S-PPO (WocaR)} & $1691$ & $1573$ & $1470$ & $1397$ & $1360$ \\
    \multicolumn{1}{l} {S-PPO (S-ATLA)} & $1935$ & $1700$ & $1456$ & $1338$ & $1217$ \\
    \multicolumn{1}{l} {S-PPO (S-PA-ATLA)} & $1883$ & $1603$ & $1438$ & $1309$ & $1176$ \\
    \multicolumn{1}{l} {S-PPO (Vanilla)} & $1959$ & $1646$ & $1447$ & $1321$ & $1206$ \\
    \multicolumn{4}{l} {\textbf{Previous smoothed agents:}} \\
    \multicolumn{1}{l} {SGLDPPO+RS} & $1773$ & $1534$ & $1464$ & $1361$ & $1212$ \\
    \multicolumn{1}{l} {RadialPPO+RS} & $2073$ & $1724$ & $1479$ & $1278$ & $1146$ \\
    \multicolumn{1}{l} {WocaRPPO+RS} & $2076$ & $1832$ & $1696$ & $1533$ & $1473$ \\
    \multicolumn{1}{l} {ATLAPPO+RS} & $1293$ & $1183$ & $1095$ & $1041$ & $966$ \\
    \multicolumn{1}{l} {PA-ATLAPPO+RS} & $1750$ & $1500$ & $1319$ & $1114$ & $1040$ \\
    \multicolumn{1}{l} {VanillaPPO+RS} & $1300$ & $1218$ & $1046$ & $970$ & $895$ \\
\bottomrule

\end{tabular*}
\end{table}
\clearpage

\subsection{Additional experiments}
\label{sec: additional experiments}

\begin{table}[H]
\caption {The reward of our S-PPO (Vanilla) under Humanoid, Ant, and Halfcheetah environments. Our S-PPO (Vanilla) outperforms the previous smoothed agents significantly without further combining other robust training algorithms. The attack budget is set to 0.075 for Humanoid and 0.15 for HalfCheetah and Ant.}
\label{table:other env sppo}
\centering
\tabcolsep=0.05cm
\footnotesize
\begin{tabular*}{\linewidth} {@{\extracolsep{\fill}} lccccc}
\toprule
    \multicolumn{1}{l}
    {Humanoid} & Clean reward & MAD attack & Min-RS attack & Optimal attack & PA-AD attack \\
    \midrule
    \multicolumn{1}{l} {S-PPO (Vanilla)} & $\bf{6956}$ & $\bf{6336}$ & $\bf{4620}$ & $\bf{6785}$ & $\bf{265}$ \\
    \multicolumn{1}{l} {VanillaPPO+RS} & $4875$ & $1581$ & $1014$ & $3350$ & $153$ \\
    \multicolumn{1}{l} {VanillaPPO} & $4913$ & $1766$ & $1040$ & $3074$ & $153$ \\
\toprule
    \multicolumn{1}{l}
    {Ant} \\
    \midrule
    \multicolumn{1}{l} {S-PPO (Vanilla)} & $5654$ & $\bf{4466}$ & $\bf{1437}$ & $\bf{871}$ & $\bf{474}$ \\
    \multicolumn{1}{l} {VanillaPPO+RS} & $6106$ & $942$ & $378$ & $-1560$ & $-1817$ \\
    \multicolumn{1}{l} {VanillaPPO} & $\bf{6141}$ & $710$ & $338$ & $-1555$ & $-1817$ \\
\toprule
    \multicolumn{1}{l}
    {Halfcheetah} \\
    \midrule
    \multicolumn{1}{l} {S-PPO (Vanilla)} & $5140$ & $\bf{4171}$ & $\bf{3577}$ & $\bf{2703}$ & $\bf{2648}$ \\
    \multicolumn{1}{l} {VanillaPPO+RS} & $5272$ & $560$ & $327$ & $-490$ & $-382$ \\
    \multicolumn{1}{l} {VanillaPPO} & $\bf{5371}$ & $527$ & $207$ & $-489$ & $-412$ \\
\bottomrule
\end{tabular*}
\end{table}
\begin{table}[H]
\caption {The reward of our S-DQN (Vanilla) with different smoothing variance $\sigma$. A higher $\sigma$ usually leads to more robust S-DQN agents but with a trade-off of decreasing the clean reward.}
\label{table:diff sigma sdqn}
\tabcolsep=0.015cm
\centering
\footnotesize
\begin{tabular*}{1\linewidth} {@{\extracolsep{\fill}} lcccccc}
\toprule
    {Pong} 
    & \multicolumn{1}{c} {Clean reward} & \multicolumn{5}{c} {S-PGD} \\
    \cmidrule(lr){3-7}
    {$\epsilon(\ell_{\infty})$} &  & $0.01$ & $0.02$ & $0.03$ & $0.04$ & $0.05$ \\
    \midrule
    {S-DQN (Vanilla) $\sigma=0.01$} & $\bf{21.0\!\pm\!0.0}$ & $8.0\!\pm\!4.0$ & $-20.8\!\pm\!0.4$ & $-20.8\!\pm\!0.4$ & $-20.8\!\pm\!0.4$ & $-20.8\!\pm\!0.4$ \\
    {S-DQN (Vanilla) $\sigma=0.05$} & $\bf{21.0\!\pm\!0.0}$ & $20.8\!\pm\!0.4$ & $\bf{20.6\!\pm\!0.5}$ & $18.6\!\pm\!2.2$ & $-11.0\!\pm\!3.4$ & $-20.6\!\pm\!0.5$ \\
    {S-DQN (Vanilla) $\sigma=0.1$} & $20.4\!\pm\!0.5$ & $\bf{21.0\!\pm\!0.0}$ & $20.4\!\pm\!0.8$ & $\bf{20.2\!\pm\!0.8}$ & $\bf{16.6\!\pm\!4.4}$ & $\bf{18.4\!\pm\!2.1}$ \\
    {S-DQN (Vanilla) $\sigma=0.15$} & $18.8\!\pm\!1.5$ & $19.6\!\pm\!1.2$ & $17.8\!\pm\!3.2$ & $17.6\!\pm\!1.9$ & $14.6\!\pm\!3.2$ & $14.4\!\pm\!3.0$ \\
\toprule
    {Freeway} \\
    \midrule
    {S-DQN (Vanilla) $\sigma=0.01$} & $\bf{34.0\!\pm\!0.0}$ & $16.6\!\pm\!1.9$ & $0.0\!\pm\!0.0$ & $0.0\!\pm\!0.0$ & $0.0\!\pm\!0.0$ & $0.0\!\pm\!0.0$ \\
    {S-DQN (Vanilla) $\sigma=0.05$} & $33.6\!\pm\!0.5$ & $\bf{33.8\!\pm\!0.4}$ & $\bf{31.6\!\pm\!1.5}$ & $6.8\!\pm\!1.7$ & $0.0\!\pm\!0.0$ & $0.0\!\pm\!0.0$ \\
    {S-DQN (Vanilla) $\sigma=0.1$} & $\bf{34.0\!\pm\!0.0}$ & $33.0\!\pm\!0.9$ & $31.4\!\pm\!1.0$ & $\bf{28.0\!\pm\!1.4}$ & $20.4\!\pm\!1.9$ & $6.6\!\pm\!2.2$ \\
    {S-DQN (Vanilla) $\sigma=0.15$} & $26.4\!\pm\!1.0$ & $26.6\!\pm\!1.6$ & $26.8\!\pm\!1.0$ & $25.2\!\pm\!1.9$ & $\bf{24.0\!\pm\!2.5}$ & $\bf{20.2\!\pm\!1.3}$ \\
\toprule
    {RoadRunner} \\
    \midrule
    {S-DQN (Vanilla) $\sigma=0.01$} &$45180\!\pm\!8944$ & $840\!\pm\!869$ & $0\!\pm\!0$ & $0\!\pm\!0$ & $0\!\pm\!0$ & $0\!\pm\!0$ \\
    {S-DQN (Vanilla) $\sigma=0.05$} & $\bf{47480\!\pm\!8807}$ & $\bf{23320\!\pm\!3932}$ & $3460\!\pm\!5924$ & $0\!\pm\!0$ & $0\!\pm\!0$ & $0\!\pm\!0$ \\
    {S-DQN (Vanilla) $\sigma=0.1$} & $39200\!\pm\!6156$ & $19640\!\pm\!2263$ & $\bf{11160\!\pm\!5644}$ & $620\!\pm\!1040$ & $0\!\pm\!0$ & $0\!\pm\!0$ \\
    {S-DQN (Vanilla) $\sigma=0.15$} & $16860\!\pm\!1334$ & $16540\!\pm\!671$ & $\bf{11160\!\pm\!993}$ & $\bf{4680\!\pm\!5629}$ & $\bf{940\!\pm\!1830}$ & $\bf{20\!\pm\!40}$ \\
\bottomrule

\end{tabular*}
\end{table}
\begin{table}[H]
\caption {The reward of our S-PPO (Vanilla) with different smoothing variance $\sigma$. The best $\sigma$ settings for Walker and Hopper are 0.2 and 0.3 respectively. However, we use $\sigma=0.2$ in every environment for simplicity.}
\label{table:diff sigma sppo}
\centering
\tabcolsep=0.05cm
\footnotesize
\begin{tabular*}{\linewidth} {@{\extracolsep{\fill}} lccccc}
\toprule
    \multicolumn{1}{l}
    {Walker} & Clean reward & MAD attack & Min-RS attack & Optimal attack & PA-AD attack \\
    \midrule
    \multicolumn{1}{l} {S-PPO (Vanilla) $\sigma=0.1$} & $\bf{4798}$ & $4316$ & $1598$ & $\bf{2853}$ & $822$ \\
    \multicolumn{1}{l} {S-PPO (Vanilla) $\sigma=0.2$} & $4552$ & $\bf{4386}$ & $\bf{3203}$ & $944$ & $\bf{1077}$ \\
    \multicolumn{1}{l} {S-PPO (Vanilla) $\sigma=0.3$} & $4207$ & $4218$ & $2098$ & $744$ & $915$ \\
\toprule
    \multicolumn{1}{l}
    {Hopper} \\
    \midrule
    \multicolumn{1}{l} {S-PPO (Vanilla) $\sigma=0.1$} & $3392$ & $2653$ & $1014$ & $569$ & $918$ \\
    \multicolumn{1}{l} {S-PPO (Vanilla) $\sigma=0.2$} & $3583$ & $2765$ & $1049$ & $995$ & $1190$ \\
    \multicolumn{1}{l} {S-PPO (Vanilla) $\sigma=0.3$} & $\bf{3642}$ & $\bf{2864}$ & $\bf{1135}$ & $\bf{1366}$ & $\bf{2083}$ \\
\bottomrule
\end{tabular*}
\end{table}
\begin{table}[H]
\caption {testing time cost and clean reward of S-DQN (Vanilla) and S-PPO (Vanilla) under different sample numbers $m$. We can see that $m=5$ is already sufficient to achieve high clean reward and the time cost is not high even with $m=100$.}
\label{table:testing time}
\centering
\tabcolsep=0.05cm
\footnotesize
\begin{tabular*}{\linewidth} {@{\extracolsep{\fill}} lcccc}
\toprule
    \multicolumn{1}{l}
    {Pong} & $m=100$ & $m=10$ & $m=5$ & $m=1$ \\
    \midrule
    \multicolumn{1}{l} {S-DQN (Vanilla) test time (sec/step)} & $0.1154$ & $0.0106$ & $0.0092$ & $0.0042$ \\
    \multicolumn{1}{l} {S-DQN (Vanilla) clean reward} & $21.0\!\pm\!0.0$ & $20.6\!\pm\!0.5$ & $20.4\!\pm\!0.5$ & $18.2\!\pm\!3.2$ \\
\toprule
    \multicolumn{1}{l}
    {Walker} \\
    \midrule
    \multicolumn{1}{l} {S-PPO (Vanilla) test time (sec/step)} & $0.0094$ & $0.0026$ & $0.0022$ & $0.0019$ \\
    \multicolumn{1}{l} {S-PPO (Vanilla) clean reward} & $4552\!\pm\!65$ & $4442\!\pm\!86$ & $4593\!\pm\!92$ & $4654\!\pm\!168$ \\
\bottomrule
\end{tabular*}
\end{table}
\begin{table}[H]
\caption {The ablation study of S-DQN (Vanilla) without Denoiser. It is hard to learn S-DQN agents without Denoiser.}
\label{table:no denoiser}
\tabcolsep=0.015cm
\centering
\footnotesize
\begin{tabular*}{1\linewidth} {@{\extracolsep{\fill}} lcccccc}
\toprule
    {Pong} 
    & \multicolumn{1}{c} {Clean reward} & \multicolumn{5}{c} {S-PGD} \\
    \cmidrule(lr){3-7}
    {$\epsilon(\ell_{\infty})$} &  & $0.01$ & $0.02$ & $0.03$ & $0.04$ & $0.05$ \\
    \midrule
    {S-DQN (Vanilla)} & $\bf{20.4\!\pm\!0.5}$ & $\bf{21.0\!\pm\!0.0}$ & $\bf{20.4\!\pm\!0.8}$ & $\bf{20.2\!\pm\!0.8}$ & $\bf{16.6\!\pm\!4.4}$ & $\bf{18.4\!\pm\!2.1}$ \\
    {S-DQN (Vanilla) w/o Denoiser} & $-21.0\!\pm\!0.0$ & $-21.0\!\pm\!0.0$ & $-21.0\!\pm\!0.0$ & $-21.0\!\pm\!0.0$ & $-21.0\!\pm\!0.0$ & $-21.0\!\pm\!0.0$ \\
\toprule
    {Freeway} \\
    \midrule
    {S-DQN (Vanilla)} & $\bf{34.0\!\pm\!0.0}$ & $\bf{33.0\!\pm\!0.9}$ & $\bf{31.4\!\pm\!1.0}$ & $\bf{28.0\!\pm\!1.4}$ & $\bf{20.4\!\pm\!1.9}$ & $\bf{6.6\!\pm\!2.2}$ \\
    {S-DQN (Vanilla) w/o Denosier} & $0.0\!\pm\!0.0$ & $0.0\!\pm\!0.0$ & $0.0\!\pm\!0.0$ & $0.0\!\pm\!0.0$ & $0.0\!\pm\!0.0$ & $0.0\!\pm\!0.0$ \\
\toprule
    {RoadRunner} \\
    \midrule
    {S-DQN (Vanilla)} & $\bf{47480\!\pm\!8807}$ & $\bf{23320\!\pm\!3932}$ & $\bf{3460\!\pm\!5924}$ & $0\!\pm\!0$ & $0\!\pm\!0$ & $0\!\pm\!0$ \\
    {S-DQN (Vanilla) w/o Denoiser} & $960\!\pm\!0$ & $0\!\pm\!0$ & $0\!\pm\!0$ & $0\!\pm\!0$ & $0\!\pm\!0$ & $0\!\pm\!0$ \\
\bottomrule

\end{tabular*}
\end{table}
\begin{table}[H]
\caption {The comparison between the smoothed attack and the non-smoothed attack for the PPO setting. We use the prefix "S-" to denote the Smoothed Attack. Unlike the DQN setting, we did not observe a significant difference between the smoothed attack and
the non-smoothed attack. }
\label{table:smoothed attack ppo}
\centering
\tabcolsep=0.05cm
\footnotesize
\begin{tabular*}{\linewidth} {@{\extracolsep{\fill}} llcccc}
\toprule
    \multicolumn{1}{l}
    {Agents} & Environments & MAD attack & Min-RS attack & Optimal attack & PA-AD attack \\
    \midrule
    \multicolumn{1}{l} {S-PPO (Vanilla)} & Walker & $\bf{4386}$ & $\bf{3203}$ & $\bf{944}$ & $\bf{1077}$ \\
    \multicolumn{1}{l} {} & Hopper & $\bf{2765}$ & $\bf{1049}$ & $995$ & $1190$ \\
\toprule
    \multicolumn{1}{l}
    {Agents} & Environments & S-MAD attack & S-Min-RS attack & S-Optimal attack & S-PA-AD attack \\
    \midrule
    \multicolumn{1}{l} {S-PPO (Vanilla)} & Walker & $4637$ & $3225$ & $949$ & $1224$ \\
    \multicolumn{1}{l} {} & Hopper & $2910$ & $1057$ & $\bf{979}$ & $\bf{1114}$ \\
\bottomrule
\end{tabular*}
\end{table}
\begin{table}[H]
\caption {Addtional results for S-DQN (SADQN) and S-DQN (WocaR). Our S-DQN can also use SADQN and WocaRDQN as base agents.}
\label{table:more sdqn results}
\tabcolsep=0.015cm
\centering
\footnotesize
\begin{tabular*}{1\linewidth} {@{\extracolsep{\fill}} lcccccc}
\toprule
    {Pong} 
    & \multicolumn{1}{c} {Clean reward} & \multicolumn{5}{c} {S-PGD} \\
    \cmidrule(lr){3-7}
    {$\epsilon(\ell_{\infty})$} &  & $0.01$ & $0.02$ & $0.03$ & $0.04$ & $0.05$ \\
    \midrule
    {S-DQN (SADQN)} & $\bf{21.0\!\pm\!0.0}$ & $\bf{21.0\!\pm\!0.0}$ & $\bf{20.6\!\pm\!0.8}$ & $20.0\!\pm\!1.1$ & $17.0\!\pm\!5.0$ & $13.4\!\pm\!2.1$ \\
    {S-DQN (WocaR)} & $19.8\!\pm\!1.6$ & $19.8\!\pm\!1.2$ & $19.0\!\pm\!2.6$ & $\bf{20.6\!\pm\!0.5}$ & $\bf{20.0\!\pm\!0.9}$ & $\bf{15.6\!\pm\!4.9}$ \\
\toprule
    {Freeway} \\
    \midrule
    {S-DQN (SADQN)} & $30.0\!\pm\!0.0$ & $30.0\!\pm\!0.0$ & $29.6\!\pm\!0.5$ & $26.4\!\pm\!1.6$ & $26.6\!\pm\!1.9$ & $26.8\!\pm\!1.0$ \\
    {S-DQN (WocaR)} & $\bf{31.2\!\pm\!1.3}$ & $\bf{31.4\!\pm\!1.0}$ & $\bf{31.8\!\pm\!1.2}$ & $\bf{30.2\!\pm\!1.2}$ & $\bf{29.6\!\pm\!1.9}$ & $\bf{28.6\!\pm\!1.4}$ \\
\toprule
    {RoadRunner} \\
    \midrule
    {S-DQN (SADQN)} & $\bf{44560\!\pm\!8724}$ & $\bf{41180\!\pm\!5618}$ & $\bf{37920\!\pm\!6478}$ & $\bf{36600\!\pm\!4994}$ & $\bf{32480\!\pm\!3803}$ & $\bf{27160\!\pm\!3287}$ \\
    {S-DQN (WocaR)} & $39120\!\pm\!6430$ & $36980\!\pm\!6978$ & $18880\!\pm\!9335$ & $7520\!\pm\!9212$ & $20\!\pm\!40$ & $2100\!\pm\!2417$ \\
\bottomrule
\end{tabular*}
\end{table}
\clearpage

\end{document}